\documentclass[10pt,twocolumn,letterpaper]{article}

\usepackage{cvpr}
\usepackage{times}
\usepackage{epsfig}
\usepackage{graphicx}
\usepackage{amsmath}
\usepackage{amssymb}

\usepackage{bm}
\usepackage{bbm}
\usepackage{float}
\usepackage{array}
\usepackage{multirow}
\usepackage{epstopdf}
\usepackage[svgnames, x11names, dvipsnames, table]{xcolor}
\usepackage{subfigure}
\usepackage{tabu}   
\usepackage{xfrac}
\usepackage{authblk}
\usepackage{stfloats}
\usepackage[final]{pdfpages}

\newcommand{\pval}[1]{\cellcolor{DodgerBlue2!25}{$p<0.001$}}
\newcommand{\pas}[1]{\cellcolor{DodgerBlue2!15}{$p=#1$}}
\newcommand{\pass}[1]{\cellcolor{DodgerBlue2!7}{$p=#1$}}
\newcommand{\fail}[1]{\cellcolor{red!15}{$p=#1$}}

\newcommand{\Sf}[0]{$S_1$}
\newcommand{\Ss}[0]{$S_2$}
\newcommand{\Tf}[0]{$T_f$}
\newcommand{\Tp}[0]{$T_p$}

\newcommand{\qsection}[1]{\vspace{5pt} \noindent \textbf{#1:}}

\makeatletter
\renewcommand\AB@affilsepx{, \protect\Affilfont}
\makeatother

\renewcommand\Affilfont{\fontsize{9}{10.8}\itshape}

\usepackage[pagebackref=true,breaklinks=true,letterpaper=true,colorlinks,bookmarks=false]{hyperref}

\cvprfinalcopy % *** Uncomment this line for the final submission

\def\cvprPaperID{8} % *** Enter the CVPR Paper ID here

\ifcvprfinal\fi
\begin{document}

%%%%%%%%% TITLE
% \title{Facial Landmark Detection in Older Adults With vs. Without Dementia}
\title{\vspace{-.8cm} Limitations and Biases in Facial Landmark Detection -- An Empirical Study on Older Adults with Dementia}

\author[1,2,3]{Azin Asgarian}
\author[3]{Shun Zhao}
\author[3,4]{Ahmed B. Ashraf}
\author[5]{M. Erin Browne}
\author[6]{Kenneth M. Prkachin}
\author[3,7,8]{\protect\\Alex Mihailidis}
\author[5,9]{Thomas Hadjistavropoulos}
\author[3,2,7,10]{Babak Taati}
\affil[1]{Georgian Partners Inc}
\affil[2]{Department of Computer Science, University of Toronto}
\affil[3]{Toronto Rehabilitation Institute, University Health Network}
\affil[4]{Department of Electrical and Computer Engineering, University of Manitoba}
\affil[5]{Department of Psychology, University of Regina}
\affil[6]{Department of Psychology, University of Northern British Columbia}
\affil[7]{Institute of Biomaterials and Biomedical Engineering, University of Toronto}
\affil[8]{Department of Occupational Science and Occupational Therapy, University of Toronto}
\affil[9]{Centre on Aging and Health, University of Regina}
\affil[10]{Vector Institute for Artiﬁcial Intelligence}

\maketitle
\thispagestyle{empty}

%%%%%%%%% ABSTRACT
\begin{abstract}
Accurate facial expression analysis is an essential step in various clinical applications that involve physical and mental health assessments of older adults (e.g. diagnosis of pain or depression). Although remarkable progress has been achieved toward developing robust facial landmark detection methods, state-of-the-art methods still face many challenges when encountering uncontrolled environments, different ranges of facial expressions, and different demographics of population. A recent study has revealed that the health status of individuals can also affect the performance of facial landmark detection methods on front views of faces. In this work, we investigate this matter in a much greater context using seven facial landmark detection methods. We perform our evaluation not only on frontal faces but also on profile faces and in various regions of the face. Our results shed light on limitations of the existing methods and challenges of applying these methods in clinical settings by indicating: 1) a significant difference between the performance of state-of-the-art when tested on the profile or frontal faces of individuals with vs. without dementia; 2) insights on the existing bias for all regions of the face; and 3) the presence of this bias despite re-training/fine-tuning with various configurations of six datasets.
\end{abstract}

\vspace{-.3cm}
\section{Introduction}
Facial landmark detection is a prerequisite for many facial analysis applications. Example clinical use cases include detecting pain in non-communicative individuals, clinical assessment of depression, and orofacial and speech assessment in individuals with a neurological motor disorder \cite{BANDINI20177, Stuhrmann2011}. For a long time, active appearance models (AAM) were the method of choice for facial landmark detection \cite{cootes2001active}. In recent years, methods beyond AAM have shown superior performance for landmark detection. Representative examples include Conditional Local Neural Fields \cite{CLNF_2013}, Coarse-to-Fine-Shape-Searching \cite{CFSS_2015}, Face Alignment Network \cite{Bulat2017}, Mnemonic Descent Method \cite{MDM2016}, and Position Map Regression Network \cite{PRNet2018}.

Despite the recent promising advances in this field, state-of-the-art methods still face many challenges when applied in realistic scenarios~\cite{biasgender, taati2019algorithmic}. To address these challenges, significant efforts have been made towards collecting images of faces in the wild (i.e. natural environment) and to cover the spectrum of age, gender, and ethnicity. However, recent studies have revealed that merely collecting more training data might not mitigate the effect of variables such as age, gender, and ethnicity~\cite{biasgender, taati2019algorithmic}. Hence, to develop algorithms that are fair with respect to potential biases, further research is required on the effect of different variables such as age, gender, and health conditions on the performance of facial landmark detection methods. 

In a recent study, Taati et al.~\cite{taati2019algorithmic} have shown that cognitive ability (healthy vs. cognitive impairment) also affects the performance of facial landmark detection methods on frontal faces of older adults. In this paper, we experimentally examine the presence of such bias in a greater scope using seven facial landmark detection methods. Moreover, we perform our evaluation on profile faces as well as on frontal faces and for various regions of the face, i.e., jaw, brows, nose, eyes, and mouth. Additionally, to further evaluate the sources of bias, we assess the performance of these methods when re-trained/fine-tuned with various training configurations of six different datasets.

Our comprehensive evaluation shows that the performance of landmark detection methods drops on the frontal and profile faces of older people with dementia as compared to cognitively healthy older adults. It also indicates that the difference in the performance between the two groups is higher in some regions of the face, such as the mouth, the eyes, and the nose, as compared to other regions such as the jaw and the brows. Moreover, our analysis shows that re-training/fine-tuning the methods improves the performance significantly on both groups, but the gap between the performance on individuals with and without dementia persists.

In the remainder of this paper, we first provide a brief overview of the datasets and landmark detection methods used in our evaluation in Sections \ref{sec:datasets} and \ref{sec:methods} respectively. Sections \ref{sec:experiments} and \ref{sec:results} describe our experimental settings and results and Section \ref{sec:conclusions} covers conclusions and future work.

%%%%%%%%%%%%%%%%%%%%%%%%%%%%%%%%%%%%%%%%%%%%%%%%%%%%%%%%%%%%%%%%%%%%%%%%%%%%%%%%%%%%%%%%%%%%%%%%%%%%%%%%%%%%%%%%%%%%%%%%%%%%%%%%%%%%%%%%%%%%%%%%%%%%%%%%%%%%%%%%%%
\section{Datasets}
\label{sec:datasets}
To conduct the experiments in this paper, we used the following six datasets: Helen~\cite{HELEN_2012}, AFW~\cite{AFW_2012}, LFPW~\cite{LFPW_2011}, MENPO Profile~\cite{zafeiriou2017menpo}, UNBC-McMaster Pain Archive~\cite{lucey2011painful}, and Pain Dataset for Dementia~\cite{hadjistavropoulos2018pain}. The MENPO Profile and Pain Dataset for Dementia include both frontal and profile faces, while the remaining four datasets only contain frontal and semi-frontal faces. The role of each dataset in each experiment (i.e. training or test) is described in \S\ref{sec:experiments}. An overview of these datasets is provided below.

\qsection{Helen} This dataset is constructed by crawling 2,330 face images from Flickr using keywords such as ``family'', ``outdoor'', ``boy'' etc. The faces were cropped and manually annotated using the PUT Face~\cite{kasinski2008put} 194 landmark points.

\qsection{Annotated Faces in the Wild (AFW)} This dataset is also collected from Flickr images and consists of 468 faces~\cite{AFW_2012}. The images of this dataset come along with annotations for six landmark points.

\qsection{Labeled Face Parts in the Wild (LFPW)} This database consists of 3,000 faces downloaded from the web using search queries (Google, Yahoo, Flickr). Annotations include 29 facial landmark points.

Since the landmark annotation for the above three datasets did not use a consistent set of points, Sagonas et~al.~\cite{SAGONAS_2013} later re-annotated a subset of examples from these datasets using a standard set of 68 landmark points\cite{MULTIPIE_2010} shown in Figure \ref{Fig:landmark_68}. From this consistently annotated subset we use 3,148 images (2,000 from Helen, 337 from AFW, and 811 from LFPW). The majority of images in these datasets are from young people and children with happy or neutral expressions. In the remainder of this paper we refer to the union of these three datasets as ``Source 1'' (\Sf). 

\qsection{MENPO Profile} This dataset contains 2,300 profile images obtained from the FDDB~\cite{fddbTech} and AFLW~\cite{koestinger2011annotated} databases and re-annotated using 39 profile view landmark points (Figure \ref{Fig:landmark_39}. We denote this dataset with $M_p$.

\qsection{UNBC-McMaster Pain Archive} The publicly available part of this dataset consists of 48,398 face images from 25 participants~\cite{UNBC_McMaster}. Participants in this dataset had a shoulder injury in one of their shoulders. During data collection, participants were asked to move their injured shoulder in one session, and their healthy shoulder in another session and their videos were recorded. Each image is annotated with the location of 68 facial landmarks, and also with the level of pain expressed in each image. Pain is coded using a 0 to 16 pain scale~\cite{prkachin2008structure} based on the Facial Action Coding System (FACS)~\cite{UNBC_McMaster}, where 0 indicates no pain and 16 indicates the highest level of pain observed. 

Using the FACS-based pain ratings, we subsampled the UNBC-McMaster dataset to 2,951 images while preserving the same distribution of pain ratings as the full dataset. In the rest of this paper we denote this subset of the UNBC-McMaster archive as ``Source 2'' (\Ss). 

\qsection{Pain Dataset for Dementia} This dataset contains data from 102 older adult participants~\cite{hadjistavropoulos2018pain} (mean age: 78.8) with and without dementia. From this dataset, Taati et~al.~\cite{taati2019algorithmic} selected data from 86 participants based on the availability of high-quality images. Of these 86 older adults, 44 were cognitively healthy and 42 were living in long-term care facilities with various degrees of dementia. Each participant was video recorded during a baseline state when lying flat on a bed, and also an exam state in which a licensed physiotherapist assisted the participant to execute a sequence of movements to identify painful areas~\cite{husebo2007mobilization}. Each session was filmed with three cameras, one capturing the frontal view and two capturing the side views (right and left). The entire dataset was annotated manually for the level of pain by trained pain coders using a FACS-based pain rating~\cite{prkachin2008structure} and a PACSLAC-II pain rating~\cite{chan2014evidence}; clinically validated methods to score pain in individuals with severe dementia~\cite{hadjistavropoulos2018pain}. 

We used two subsets of this data in our experiments which we denote by ``Target:Frontal'' (\Tf) and ``Target:Profile'' (\Tp).  To construct ``Target:Frontal'', we sub-sampled 688 frontal view images from the 86 participants. To ensure the existence of expressions corresponding to various levels of pain for each person, images of the exam state were clustered into 7 groups based on the level of pain expressed and one image was chosen at random from each group. Also, to account for the existence of neutral expressions, one image from each participant was selected at random from the baseline state. All images were manually rotated when needed to place the face in an upright position and were manually annotated using the standard 68 landmark points. Similarly, to build ``Target:Profile'', 679 profile view images were sub-sampled and manually annotated using the 39 landmark points shown in Figure \ref{Fig:landmark_39}.

%%%%%%%%%%%%%%%%%%%%%%%%%%%%%%%%%%%%%%%%%%%%%%%%%%%%%%%%%%%%%%%%%%%%%%%%%%%%%%%%%%%%%%%%%%%%%%%%%%%%%%%%%%%%%%%%%%%%%%%%%%%%%%%%%%%%%%%%%%%%%%%%%%%%%%%%%%%%%%%%%%
\section{Landmark Detection Methods}
\label{sec:methods}
\label{SubSec:Methods}
The following methods (and models) were used in our analysis: Active Appearance Models (AAM)~\cite{cootes1998active}, Constrained Local Neural Field (CLNF)~\cite{CLNF_2013}, Coarse-to-Fine Shape Searching (CFSS)~\cite{CFSS_2015}, Face Alignment Network (FAN)~\cite{Bulat2017}, Mnemonic Descent Method (MDM)~\cite{MDM2016}, and Position Map Regression Network (PRNet)~\cite{PRNet2018}. In the following we briefly review these methods.

\qsection{Active Appearance Models (AAM)} 
An AAM~\cite{cootes1998active} is a generative model that captures variations of shape and appearance of a deformable object from a set of labeled images. The model thus has two components, one for shape, and another for appearance. To train the AAM model, first Procrustes analysis is applied on training data and then PCA is performed on the shape labels and image pixels, to build the shape and appearance models. During fitting, the AAM initializes from the mean shape and tries to find the best set of parameters that minimizes the difference between the input image and the reconstructed image (based on shape and appearance parameters).

\vspace{-.05cm}
\qsection{Constrained Local Neural Field (CLNF)} 
A CLNF by Baltrušaitis et~al.~\cite{CLNF_2013} is an instance of the Constrained Local Model (CLM)~\cite{CLM} that incorporates Local Neural Field patch experts. Local Neural Field patch experts are applied on the landmark areas to learn non-linear relationships of the pixels around the landmark. Similar to AAM, the CLNF also has a shape component that models the location of the landmark points as a combination of a mean shape and a set of transformations. During fitting, the CLNF model tries to find the best set of transformation parameters that optimizes the patch expert responses while taking the reliability of each patch expert into account. 
\vspace{-.05cm}

\qsection{Coarse-to-Fine Shape Searching (CFSS)} 
Unlike many facial landmark detection methods that require an initial shape (usually the mean shape) to start the fitting process, Zhu et~al.~\cite{CFSS_2015} proposed CFSS which initializes searching from a shape space. A CFSS builds a large space of candidate shapes and performs face alignment in a given number of stages. The model starts searching by sampling from a large region in the shape space and estimates a finer sub-region to perform searches in the later stages. The adaptive stage-by-stage approach prevents the model from being trapped in local optima due to poor initialization.
\vspace{-.05cm}

\qsection{Face Alignment Network (FAN)} 
The FAN model, proposed by Bulat et~al.~\cite{Bulat2017}, regresses landmark heatmaps directly. To regress the 2D landmarks, FAN-2D employs a stack of four hourglass (HG) networks~\cite{newell2016stacked} and trains them with RGB images as input, and 68 2D Gaussian heatmaps as target output, one for each of the 68 facial landmark points. A FAN-3D network is jointly trained with an additional 2D-to-3D FAN network, where FAN-3D predicts the 2D projection of the 3D landmark points and the 2D-to-3D FAN estimates the corresponding z coordinates for the 2D landmark points predicted by FAN-3D. In this work we fine-tuned FAN-2D with everything but the last hourglass network frozen, which we refer to as FFAN-HG.

\qsection{Mnemonic Descent Method (MDM)} 
Trigeorgis et~al. proposed MDM~\cite{MDM2016}, which is an end-to-end face alignment model; i.e., it predicts the landmark coordinates directly from raw image pixels. Instead of more traditional hand-crafted features such as HOG or SIFT, MDM learns a two layer Convolutional Neural Network (CNN) as the feature extractor. For fitting, the MDM model employs the idea of learning descent directions~\cite{SDM_2013} with an additional RNN component that learns information about the past descent directions during training and then uses this information in the fitting process. 

\qsection{Position Map Regression Network (PRNet)} 
The PRNet model proposed by Feng et~al.~\cite{PRNet2018} employs a Neural Network architecture that contains Residual Blocks~\cite{ResidualBlock} and convolutional layers to simultaneously reconstruct the 3D facial structure and perform facial landmark alignment. For training, ground truth 3D facial shapes are first projected into UV space (a 2D image representation of 3D coordinates) and then the obtained UV images are used to train the model. For fitting, the PRNet model first predicts the UV images from the input image and then 3D facial structure and aligned facial landmarks are derived from the predicted UV images.

%%%%%%%%%%%%%%%%%%%%%%%%%%%%%%%%%%%%%%%%%%%%%%%%%%%%%%%%%%%%%%%%%%%%%%%%%%%%%%%%%%%%%%%%%%%%%%%%%%%%%%%%%%%%%%%%%%%%%%%%%%%%%%%%%%%%%%%%%%%%%%%%%%%%%%%%%%%%%%%%%%
\section{Experiments}
\label{sec:experiments}
For fair and comprehensive evaluation, we consider four different experiments. In the first three experiments we compare the performance of all methods on the cognitively healthy older adult subset ($44\times8=352$ images) vs. the dementia subset ($42\times8=336$ images) of the \Tf\thinspace dataset. In the last experiment, we use the healthy subset (338 images) and the dementia subset (325 images) of the \Tp\thinspace dataset for evaluation. The training set configurations explored in each experiment are described below. In any configuration that involved training examples from \Tf\thinspace and \Tp,  leave-one-participant-out cross-validation was employed to ensure images from the same person did not appear in both training and test data. 

\subsection{Experimental Settings}
\vspace{-.1cm}
\qsection{Experiment 1} 
In this experiment, we used the off-the-shelf versions of the following seven methods: CLNF~\cite{CLNF_2013}, CFSS~\cite{CFSS_2015}, AAM~\cite{alabort2014menpo}, FAN-2D and FAN-3D~\cite{Bulat2017}, MDM~\cite{MDM2016}, and PRNet~\cite{PRNet2018}. Many groups offer pre-trained AAM models which are usually trained on \Sf. For consistency, we also trained the the AAM model on \Sf. 

\qsection{Experiment 2} 
In the second experiment, we evaluated different models when re-trained/fine-tuned with \Tf. Models AAM, CLNF, and CFSS were re-trained with \Tf. However, since \Tf\thinspace was significantly smaller than the original dataset used to train model FAN-2D, a fine-tuned version of this model with \Tf\thinspace (which we call FFAN-HG) was included in this experiment. Models MDM, FAN-3D, and PRNet were excluded due to unavailability of the training code and lack of 3D ground truth landmark annotations for images in \Tf.

\qsection{Experiment 3} 
In our third experiment, we evaluated various methods when re-trained with the following configurations: $S_1, S_2, S_1 \cup S_2, T_f \cup S_1, \; T_f \cup S_2, \; T_f \cup S_1 \cup S_2$. In addition to methods MDM, FAN-3D, and PRNet, method FAN-2D was also excluded from this experiment as it is originally trained on a super set of \Sf. 

\qsection{Experiment 4} 
The off-the-shelf versions of methods FAN-2D, FAN-3D, and PRNet work on profile faces; therefore, they were included in this experiment. However, the rest of the methods only work on frontal and semi-frontal faces and need re-training to handle profile faces. We re-trained model AAM with configurations $T_p, M_p, T_p \cup M_p$. But considering the size of these training configurations, re-training was not an option for the rest of the methods and thus they were excluded from this experiment. 

\subsection{Evaluation} 
We compare the performance of different methods on the cognitively healthy older adult subset (H) versus the dementia subset (D) of \Tf\thinspace and \Tp\thinspace in terms of the convergence rate. To measure the convergence rate, we use its standard definition in the literature~\cite{asgarian2017subspace, Bulat2017, MDM2016} as the percentage of test examples that converge to the ground truth landmark points given a tolerance in the root mean squared (RMS) fitting error (here, 5\% of the face diagonal). 

We also show convergence curves that plot the percentage of test examples converged to the ground truth as a function of tolerance in RMS fitting error (normalized by the face diagonal). A typical comparison point is the point on the curve corresponding to 5\% tolerance. We perform this evaluation for the landmark points spanning the whole face and also for points that lie in specific regions i.e., jaw, brows, nose, eyes, and mouth. To evaluate statistical significance, we use the non-parametric Wilcoxon rank-sum test (on RMS errors) and consider three standard significance levels 0.05, 0.01 and 0.001. 

\begin{figure}[ht!]
\vspace{-0.2cm}
\centering    
    \subfigure[68 landmark points (frontal)]{\label{Fig:landmark_68}\includegraphics[width=.22\textwidth]{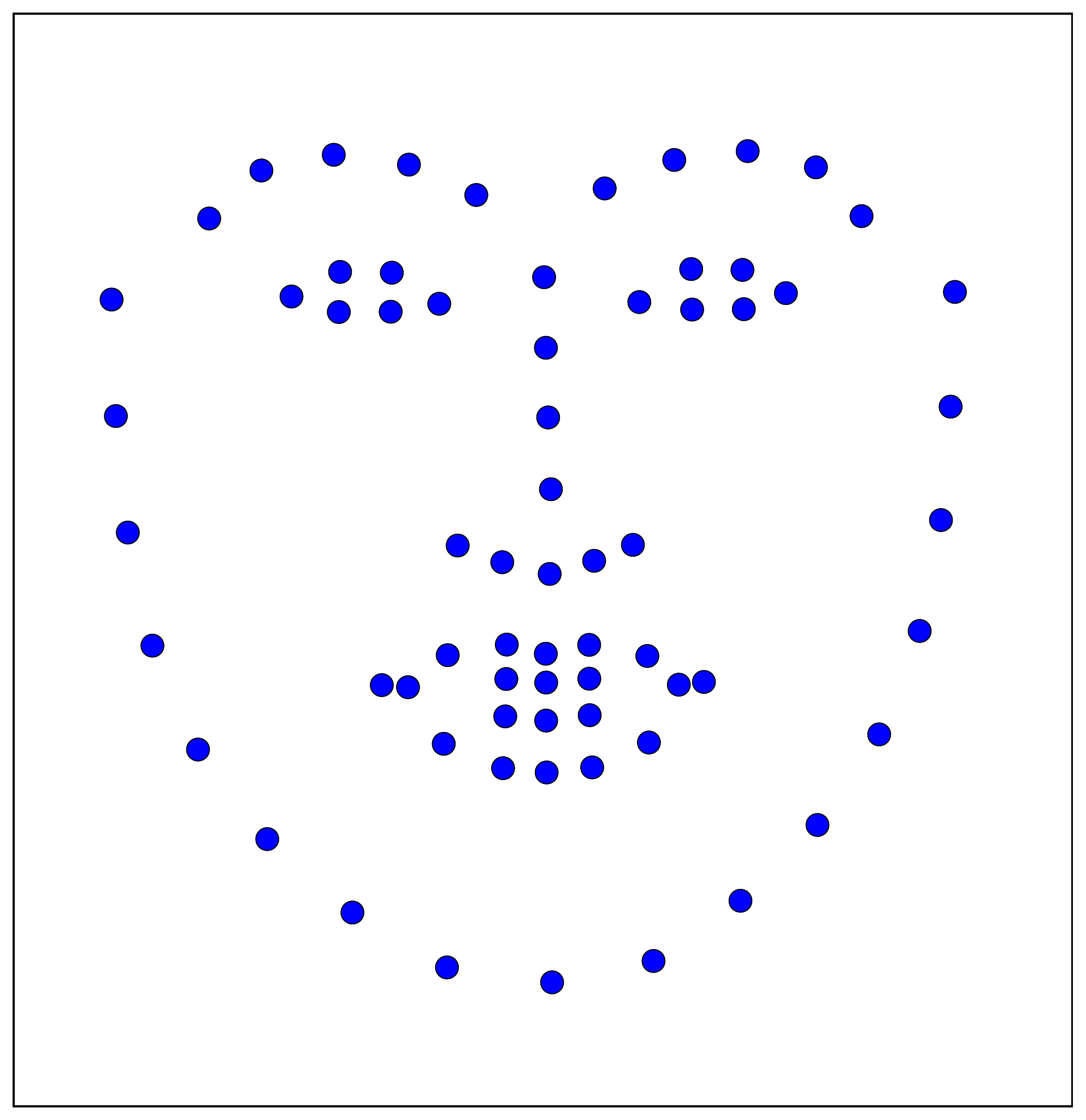}} \quad
    \subfigure[39 landmark points (profile)] {\label{Fig:landmark_39}\includegraphics[width=.22\textwidth]{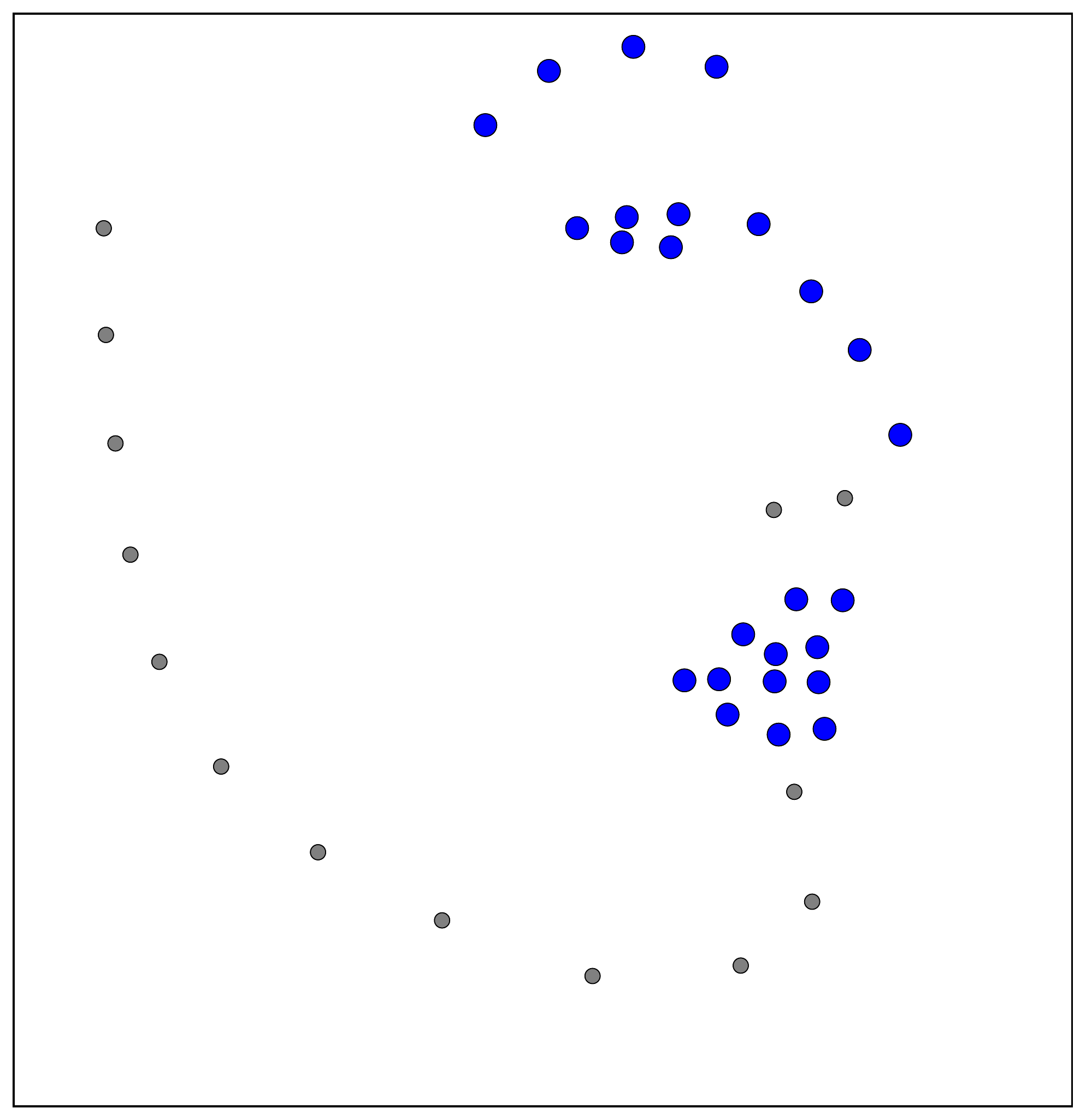}}
\caption{Different sets of landmark points used in the evaluations.}
\label{Fig:Landmark}
\vspace{-.1cm}
\end{figure}

To ensure a fair comparison, for each image in \Tf\thinspace and \Tp, the same face bounding box (detected by the Dlib face detector~\cite{king2009dlib}) was used to initialize all the models. Results in Experiments 1-3 are evaluated using the 68 landmark points. In Experiment 4, the AAM model gives 39 landmark points while the rest of the methods output the standard set of 68 landmark points~\cite{MULTIPIE_2010}. The two sets of landmark points are shown in Figures \ref{Fig:landmark_39} and \ref{Fig:landmark_68}. Since there is not a one to one correspondence between all the points in these mark-ups, results in Table \ref{Tab:Expt4} and Figure \ref{Fig:Expt4} are evaluated on the 25 points that are common between the two mark-ups from all regions of the face except the jaw line. These 25 points are shown in blue in Figure \ref{Fig:landmark_39}. 

\section{Results}
\label{sec:results}
The convergence rates obtained for all regions of the face with the methods explored in experimental settings 1, 2, and 4 on healthy (H) and dementia (D) subsets of \Tf\thinspace and \Tp\thinspace are shown respectively in Tables \ref{Tab:Expt1}, \ref{Tab:Expt2}, and \ref{Tab:Expt4}. The results of Wilcoxon rank-sum tests that evaluate the statistical significance of difference between the performance on healthy (H) and dementia (D) subsets of \Tf\thinspace and \Tp\thinspace are also reported for all methods and regions of the face.

\begin{figure*}[t]
\centering
\subfigure{\includegraphics[width=.32\textwidth]{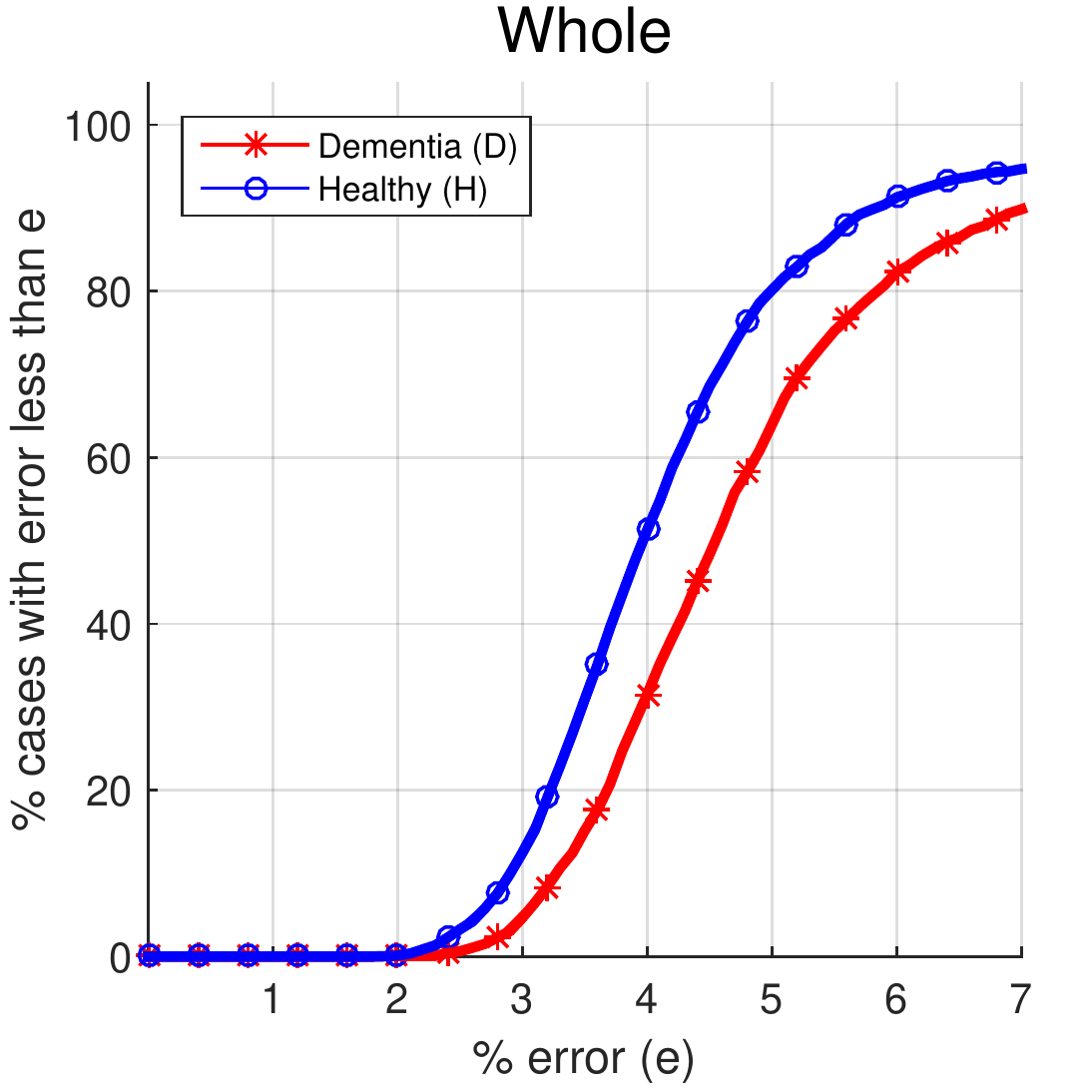}}
\subfigure{\includegraphics[width=.32\textwidth]{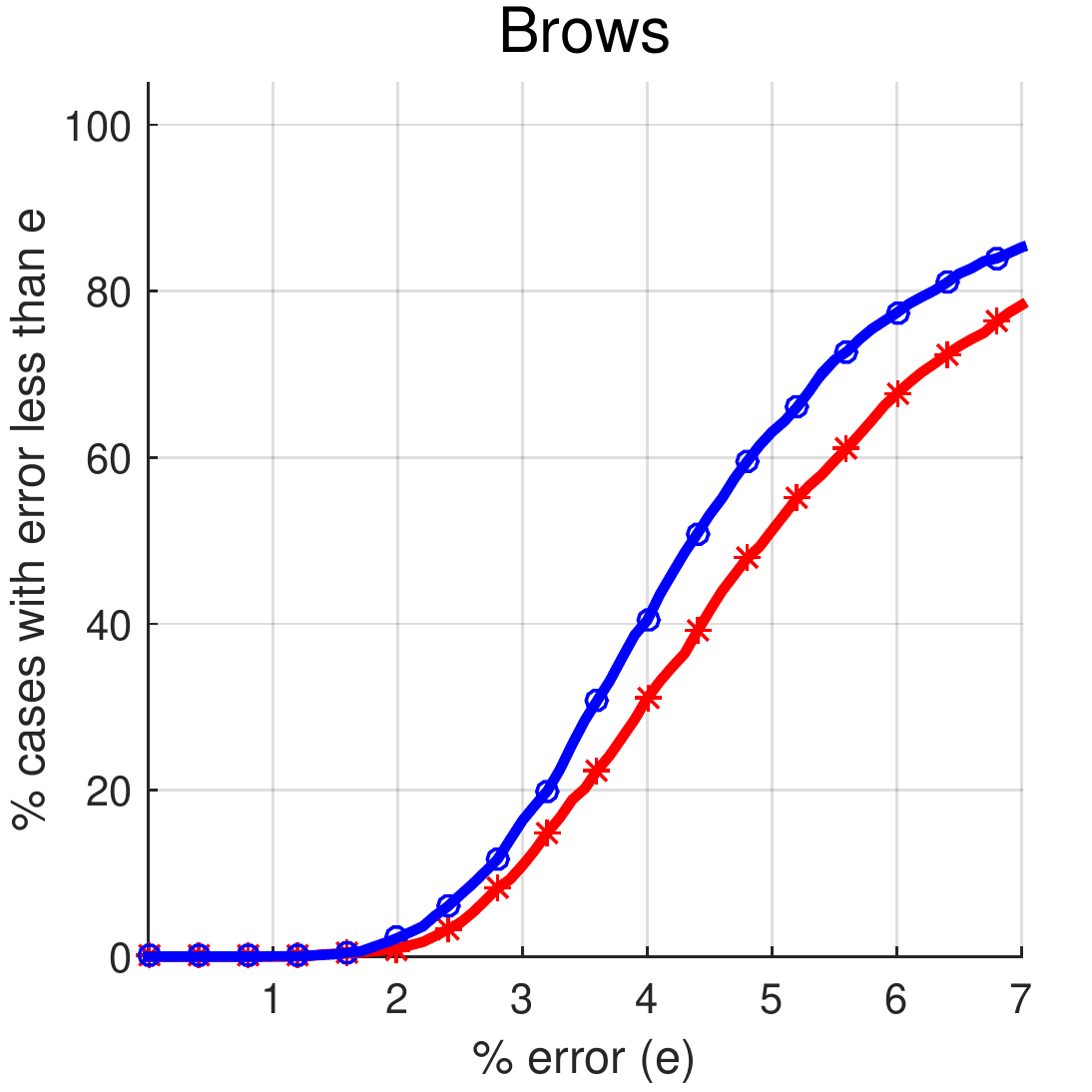}}
\subfigure{\includegraphics[width=.32\textwidth]{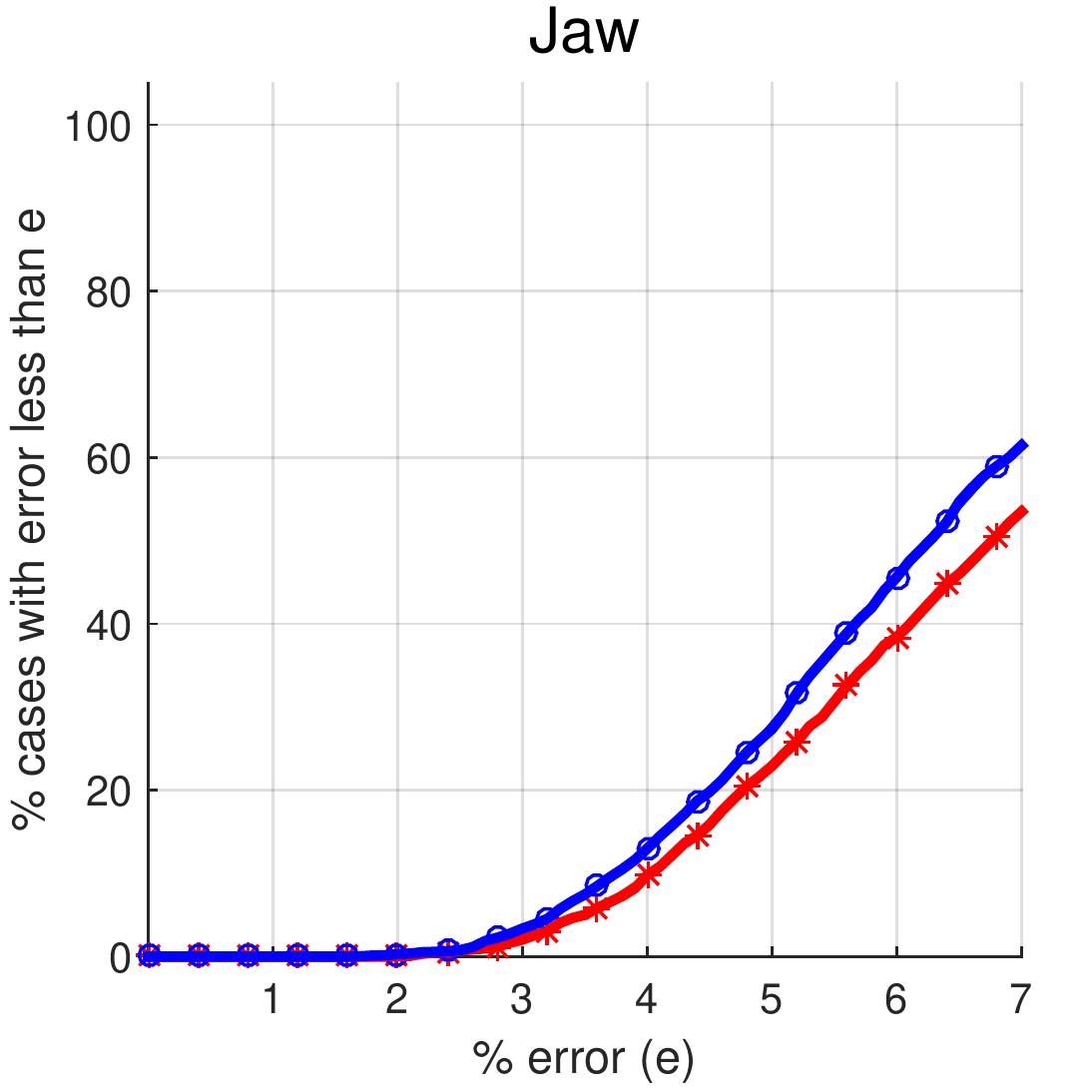}}
\subfigure{\includegraphics[width=.32\textwidth]{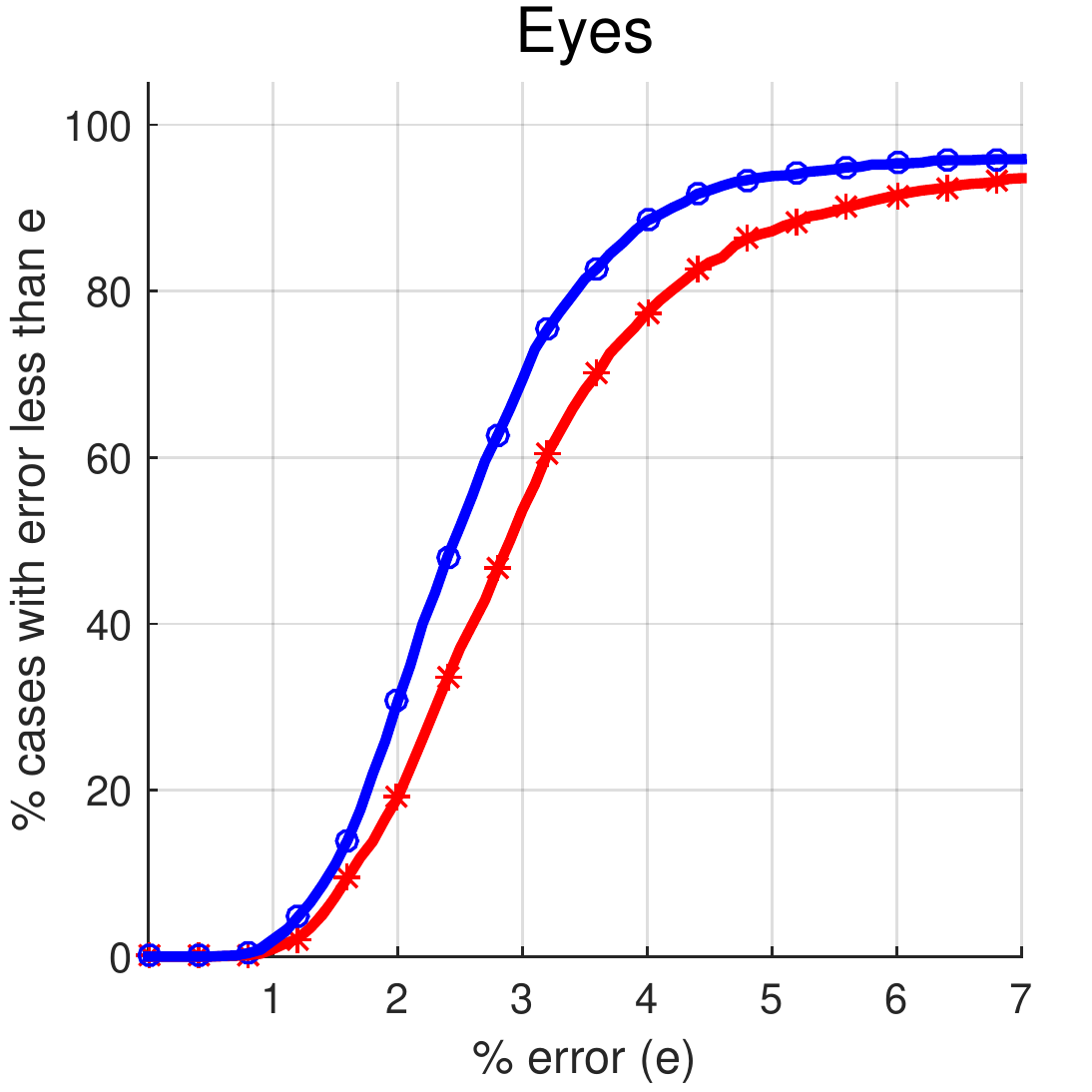}}
\subfigure{\includegraphics[width=.32\textwidth]{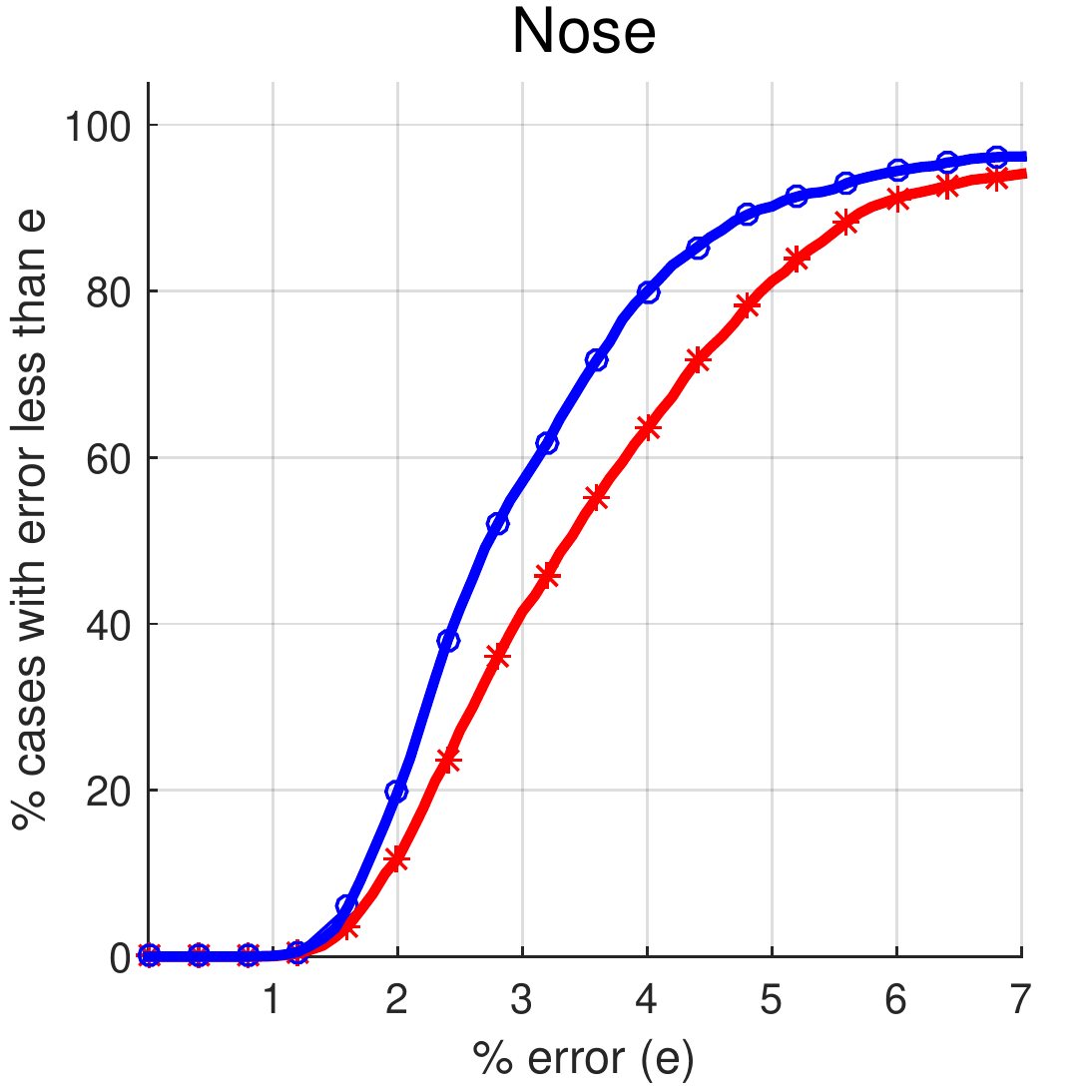}}
\subfigure{\includegraphics[width=.32\textwidth]{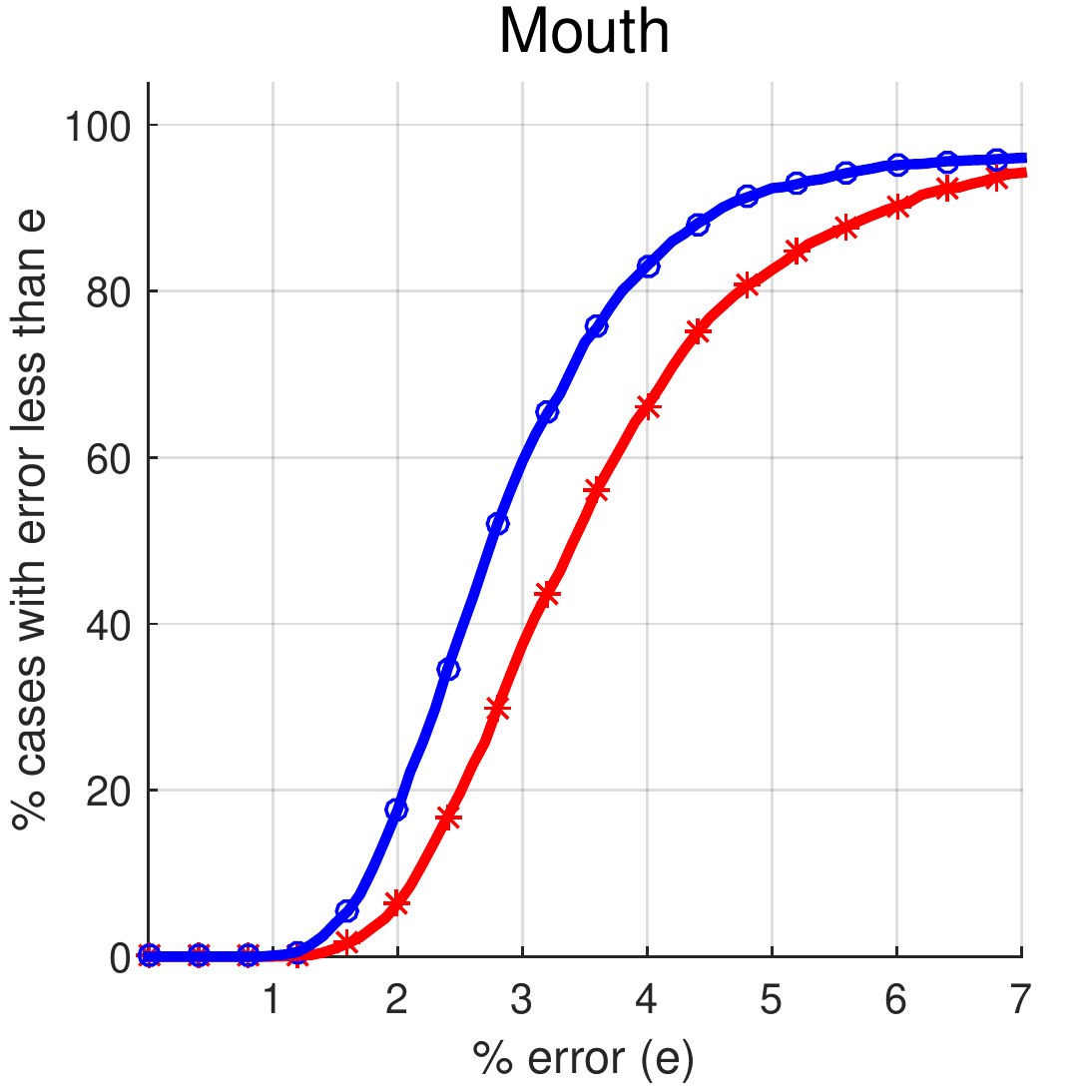}}
\caption{Experiment 1: Comparison of the average convergence curves obtained on healthy subset (H) and dementia subset (D) of \Tf. The values on y-axis are averaged over seven methods: CLNF, CFSS, AAM, FAN-2D, FAN-3D, MDM, and PRNet. }
\label{Fig:Expt1}
\vspace{.3cm}
\end{figure*}

\begin{figure*}[t]
\vspace{-.4cm}
\centering
\subfigure{\includegraphics[width=.32\textwidth]{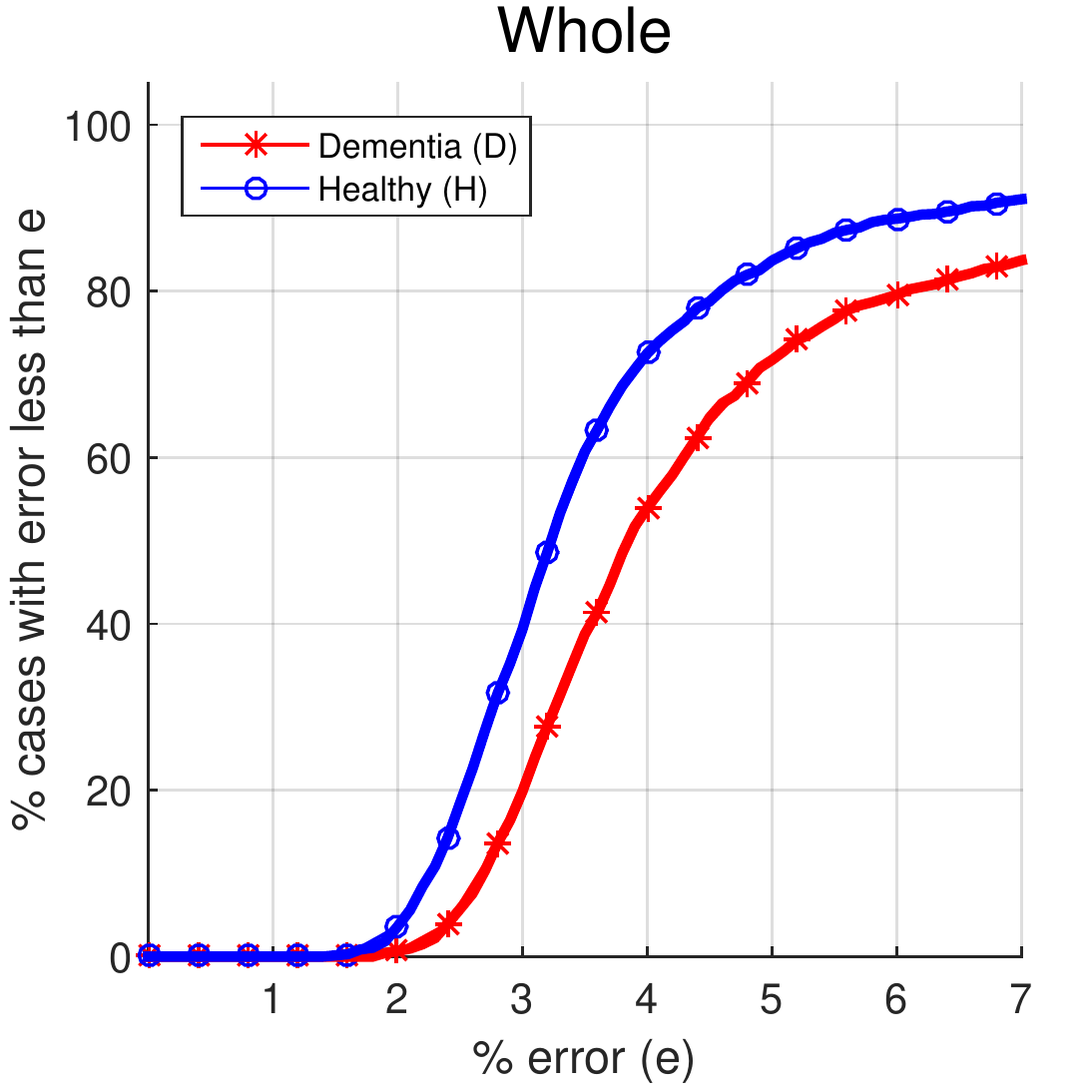}}\hspace{1mm}
\subfigure{\includegraphics[width=.32\textwidth]{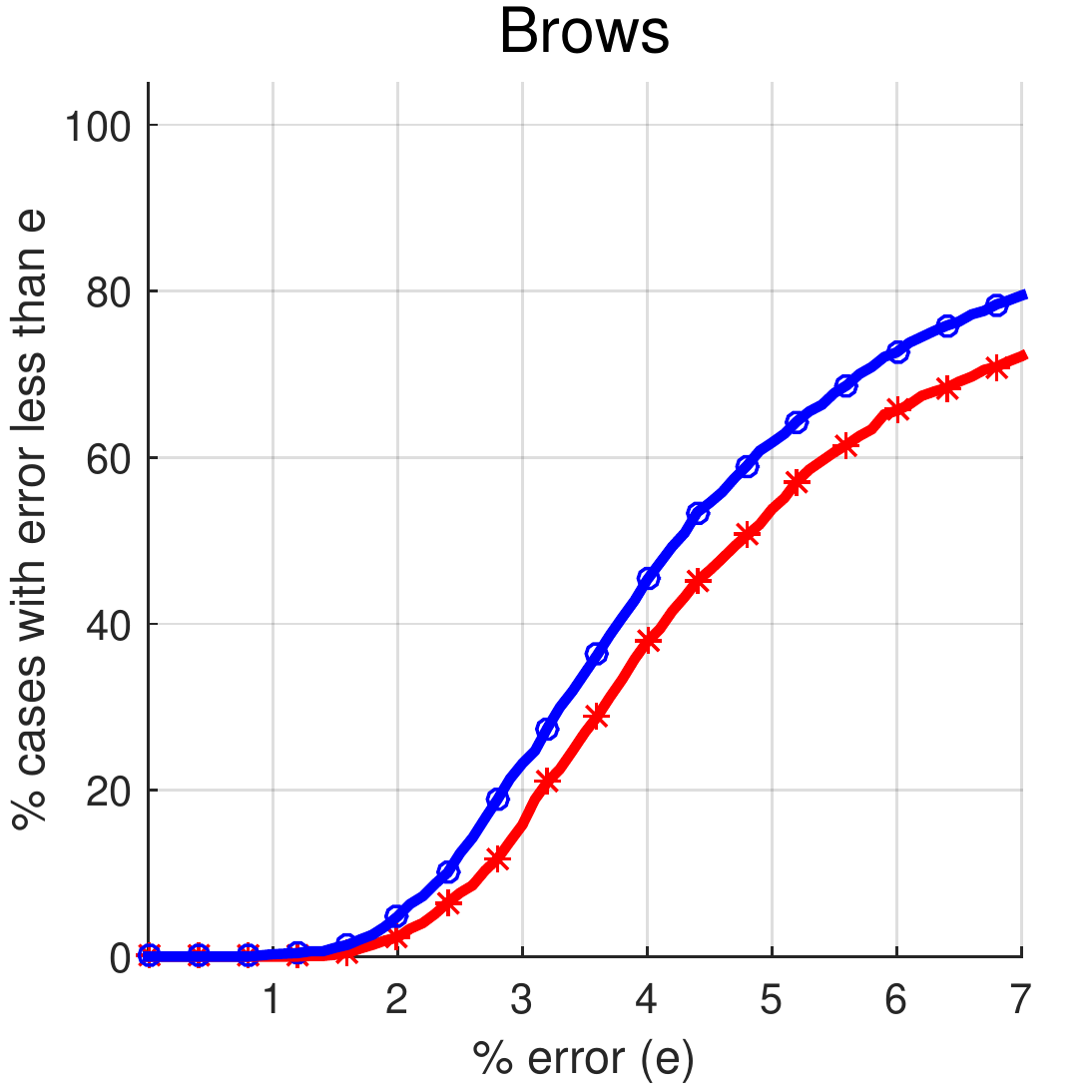}}\hspace{1mm}
\subfigure{\includegraphics[width=.32\textwidth]{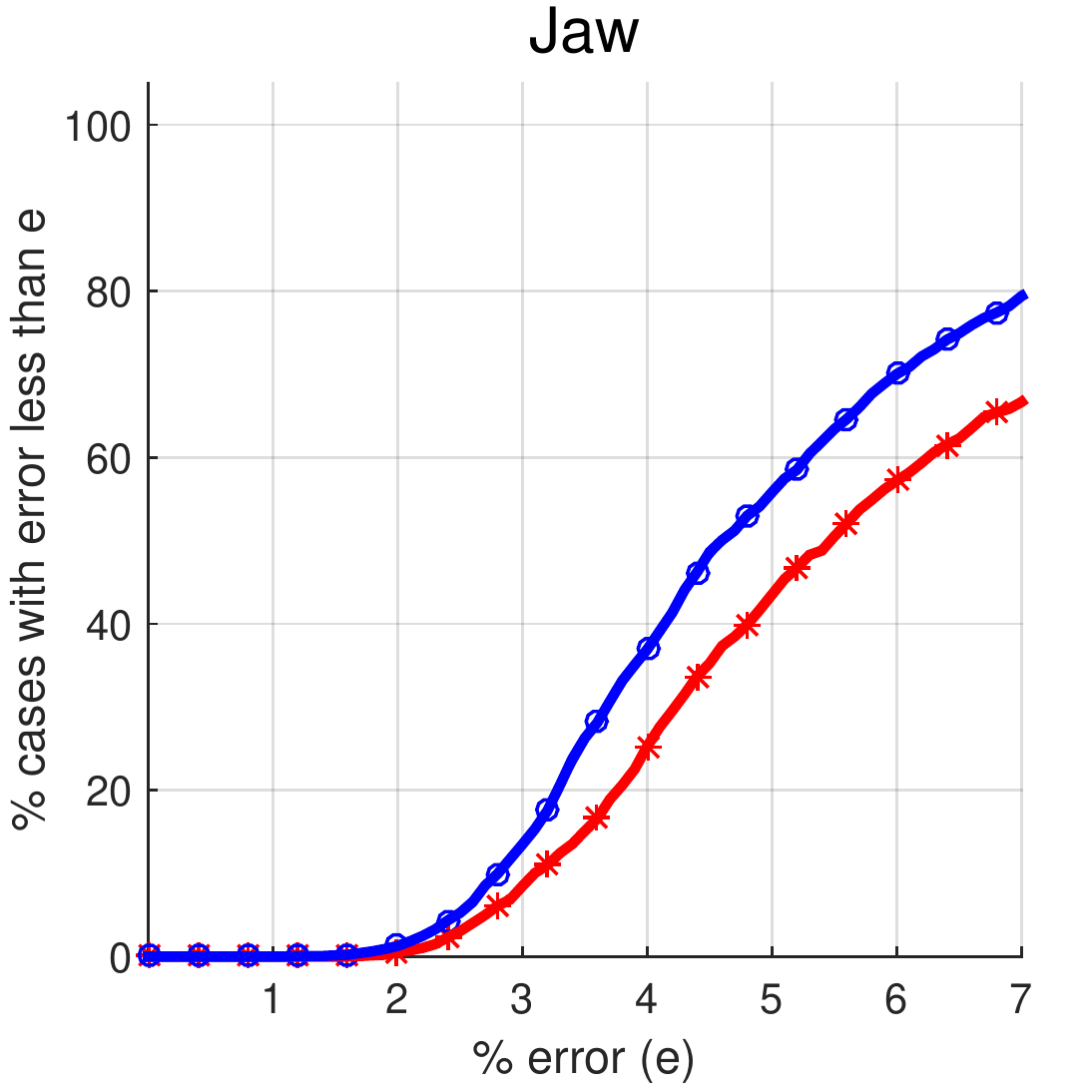}}\hspace{1mm}
\subfigure{\includegraphics[width=.32\textwidth]{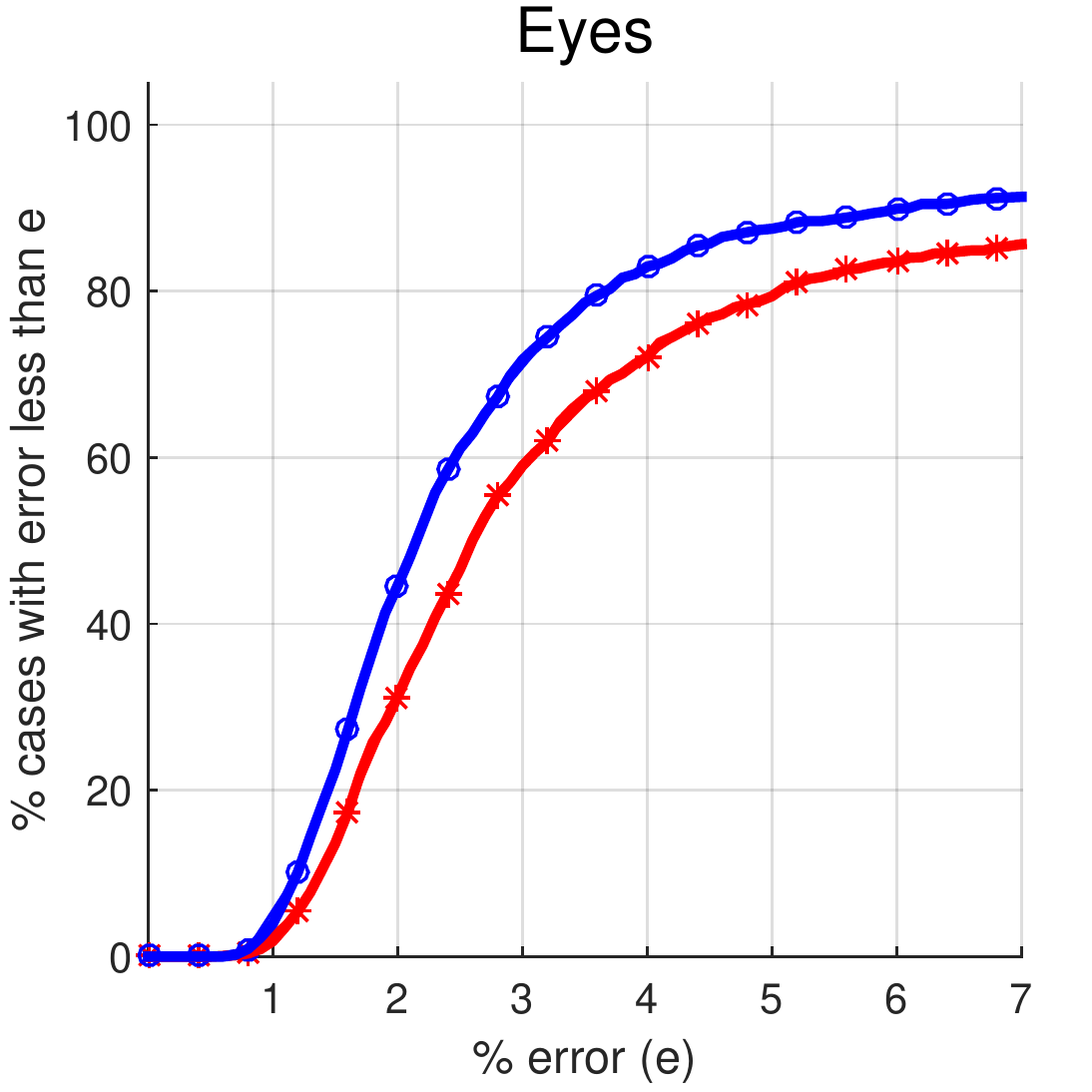}}\hspace{1mm} 
\subfigure{\includegraphics[width=.32\textwidth]{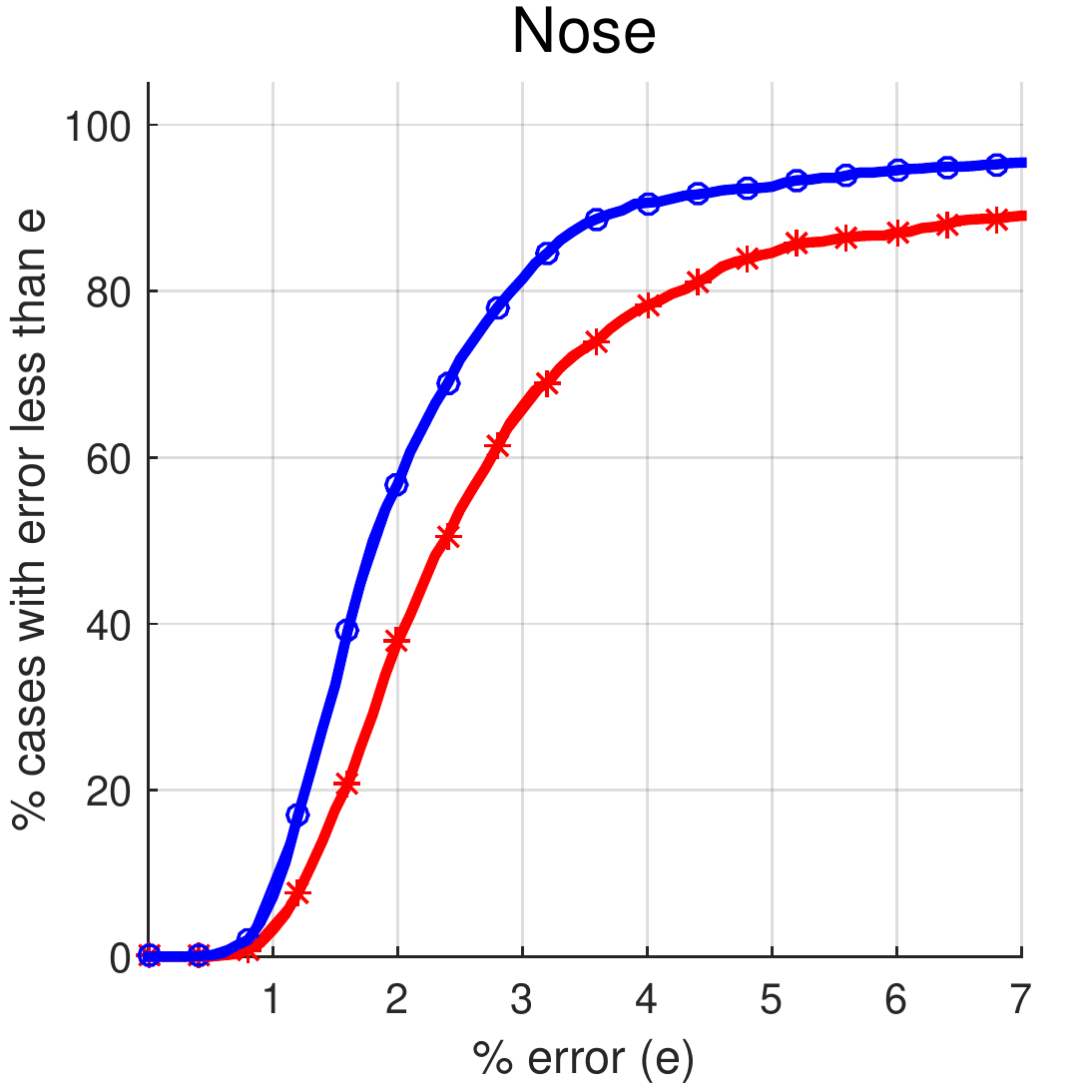}}\hspace{1mm} 
\subfigure{\includegraphics[width=.32\textwidth]{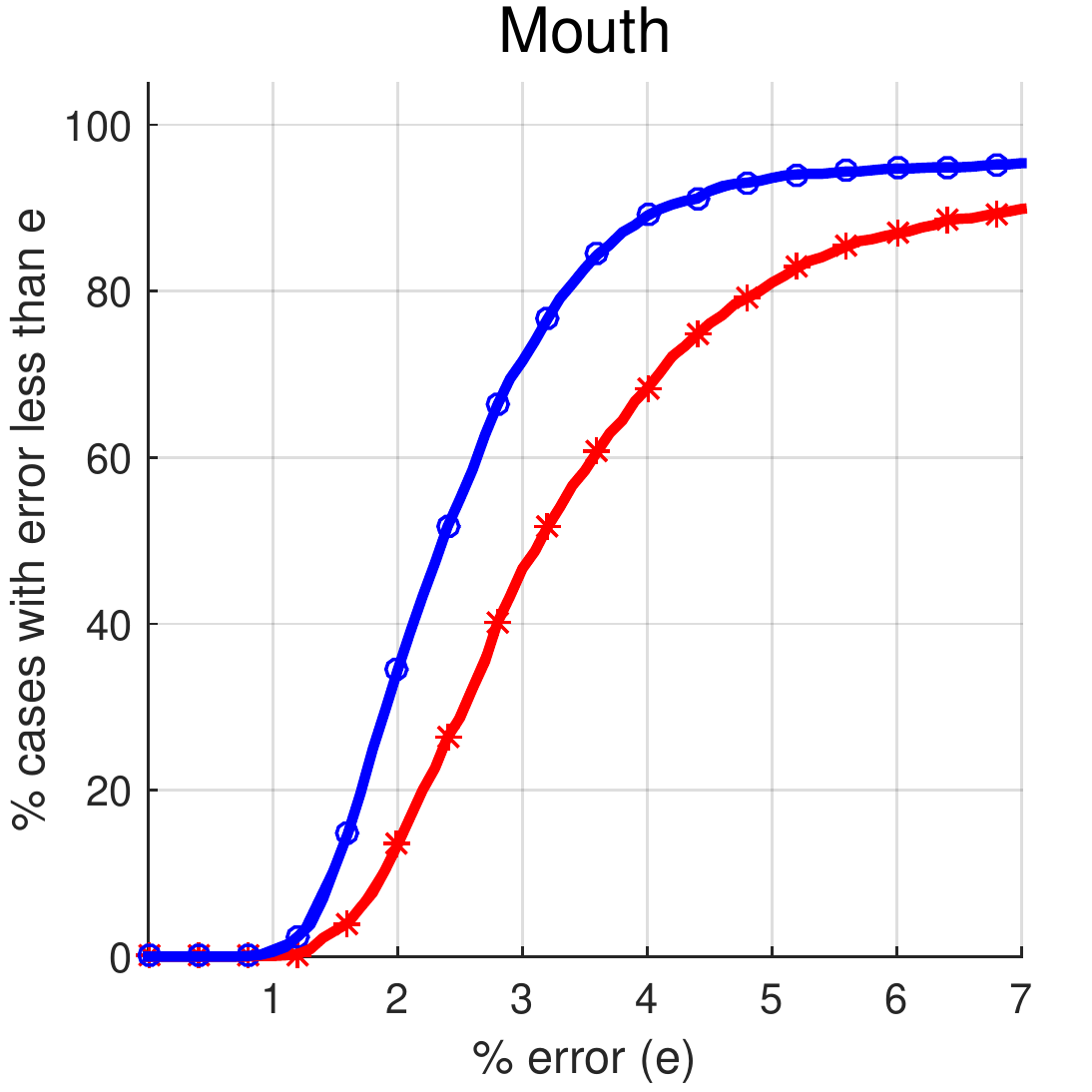}}\hspace{1mm} 
\caption{Experiment 2: Comparison of the average convergence curves obtained on healthy subset (H) and dementia subset (D) of \Tf. The values on y-axis are averaged over four methods: CLNF, CFSS, AAM, and FFAN-HG.}
\label{Fig:Expt2}
\end{figure*}

\begin{figure*}[!ht]
\centering
\subfigure{\includegraphics[width=.32\textwidth]{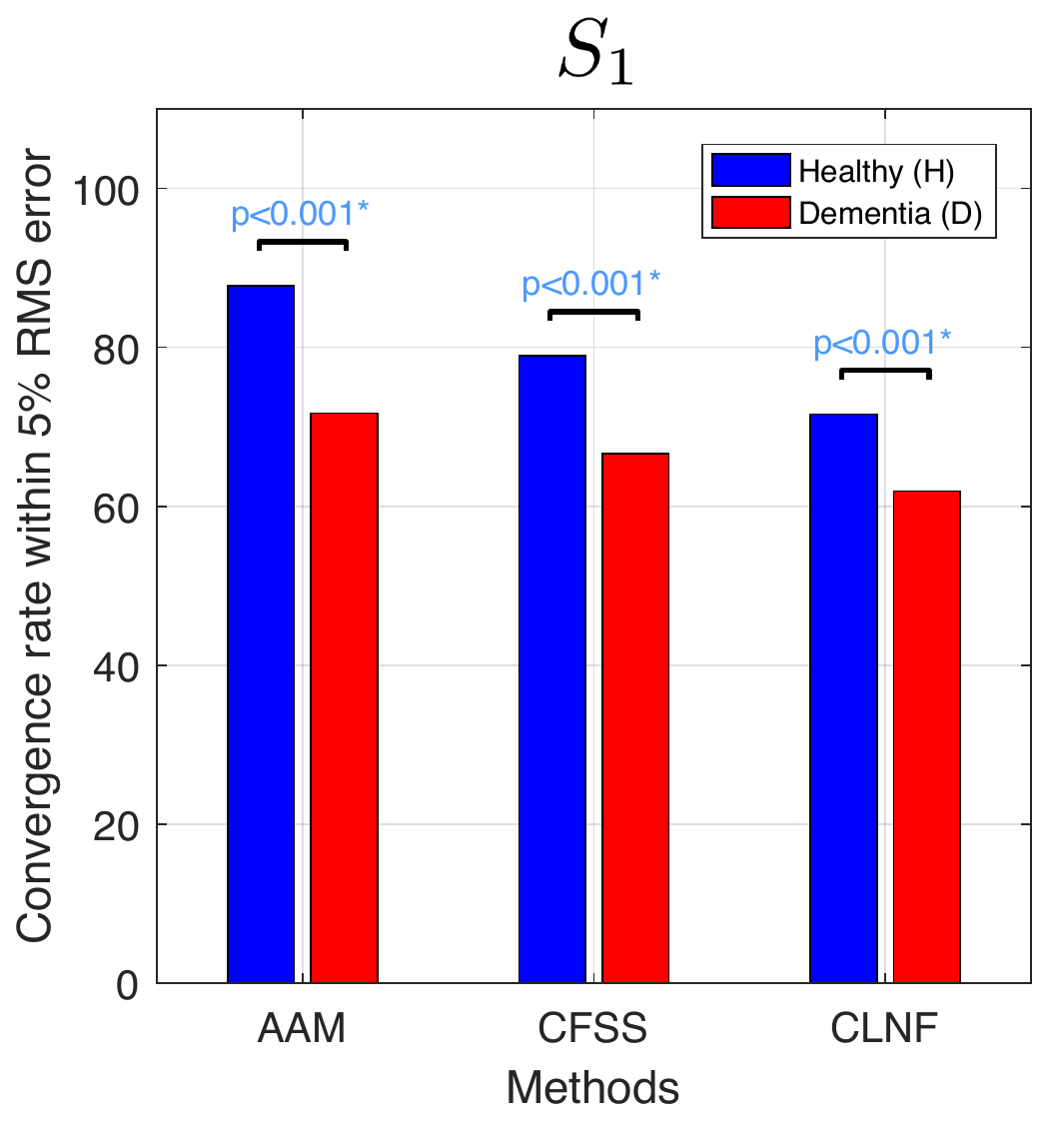}}\hspace{1mm}
\subfigure{\includegraphics[width=.32\textwidth]{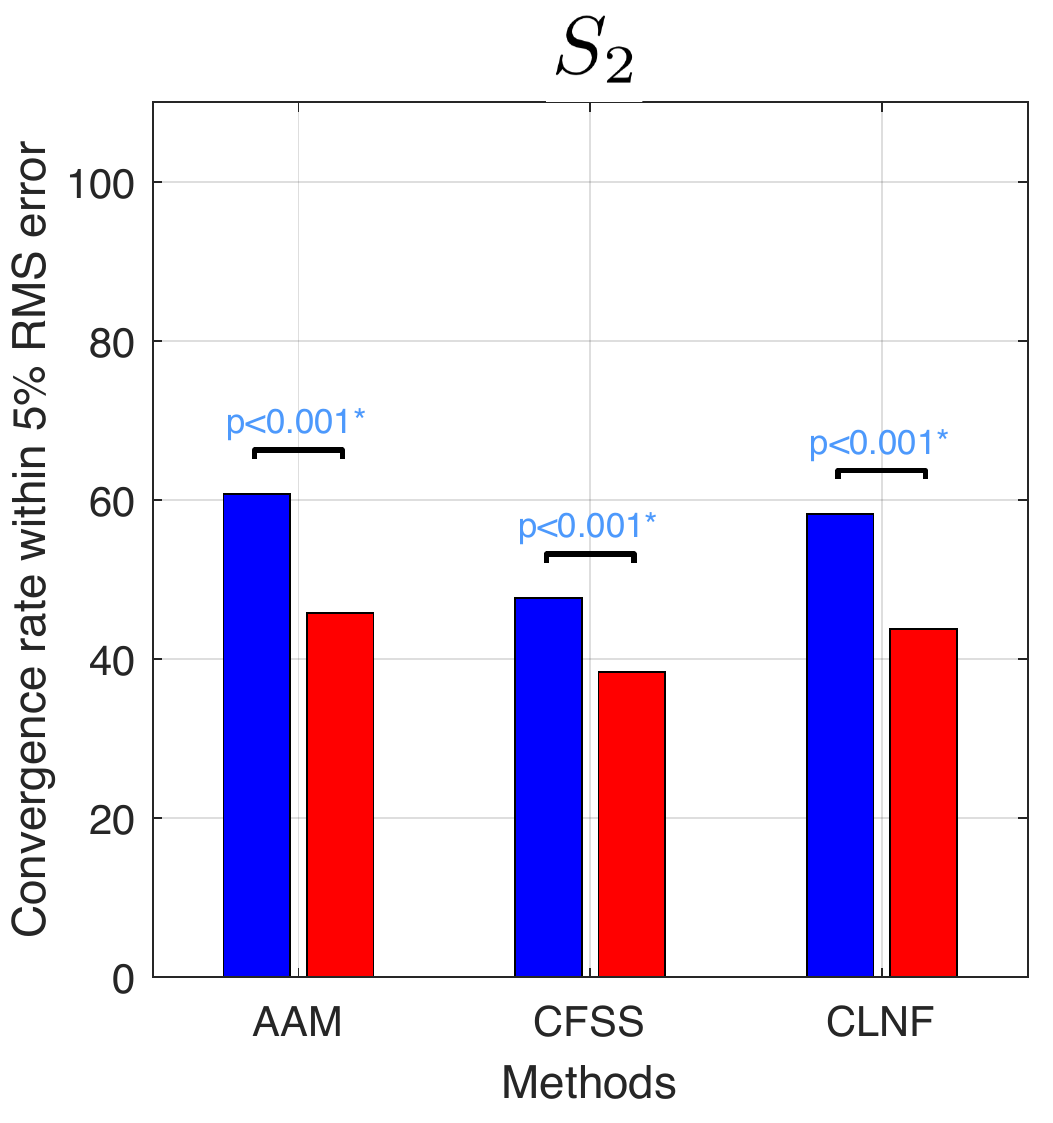}}\hspace{1mm}
\subfigure{\includegraphics[width=.32\textwidth]{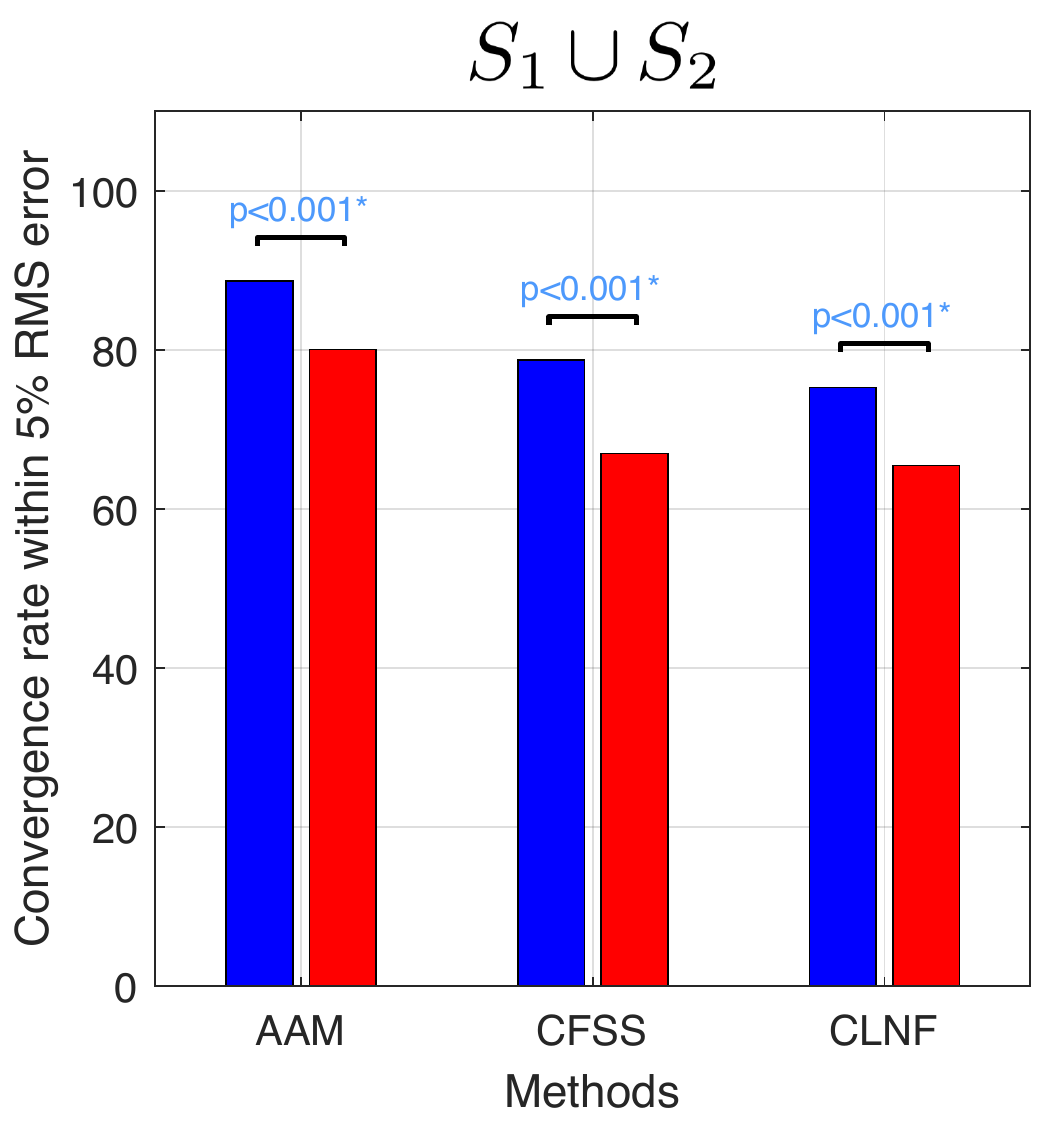}}\hspace{1mm}
\subfigure{\includegraphics[width=.32\textwidth]{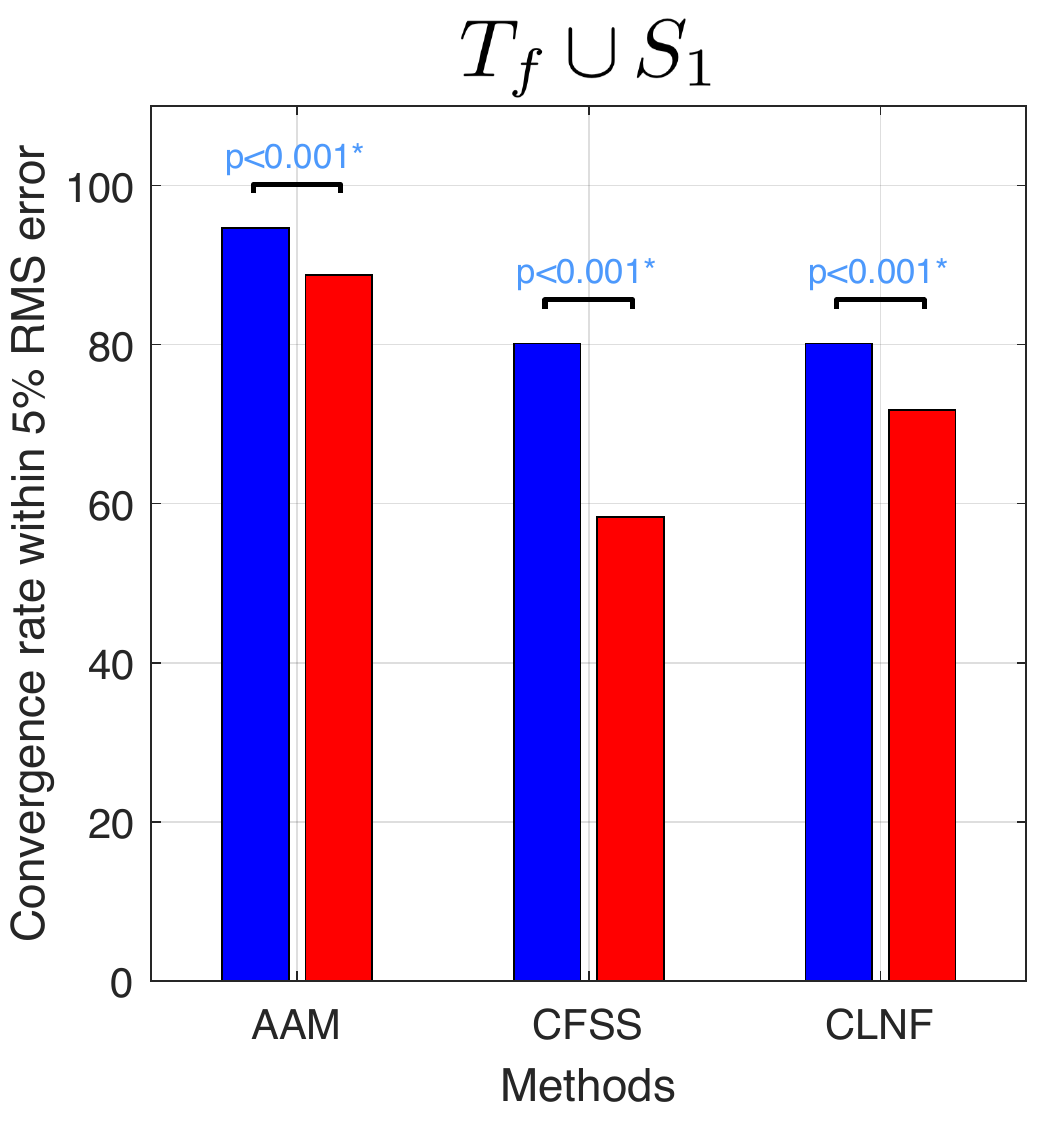}}\hspace{1mm} 
\subfigure{\includegraphics[width=.32\textwidth]{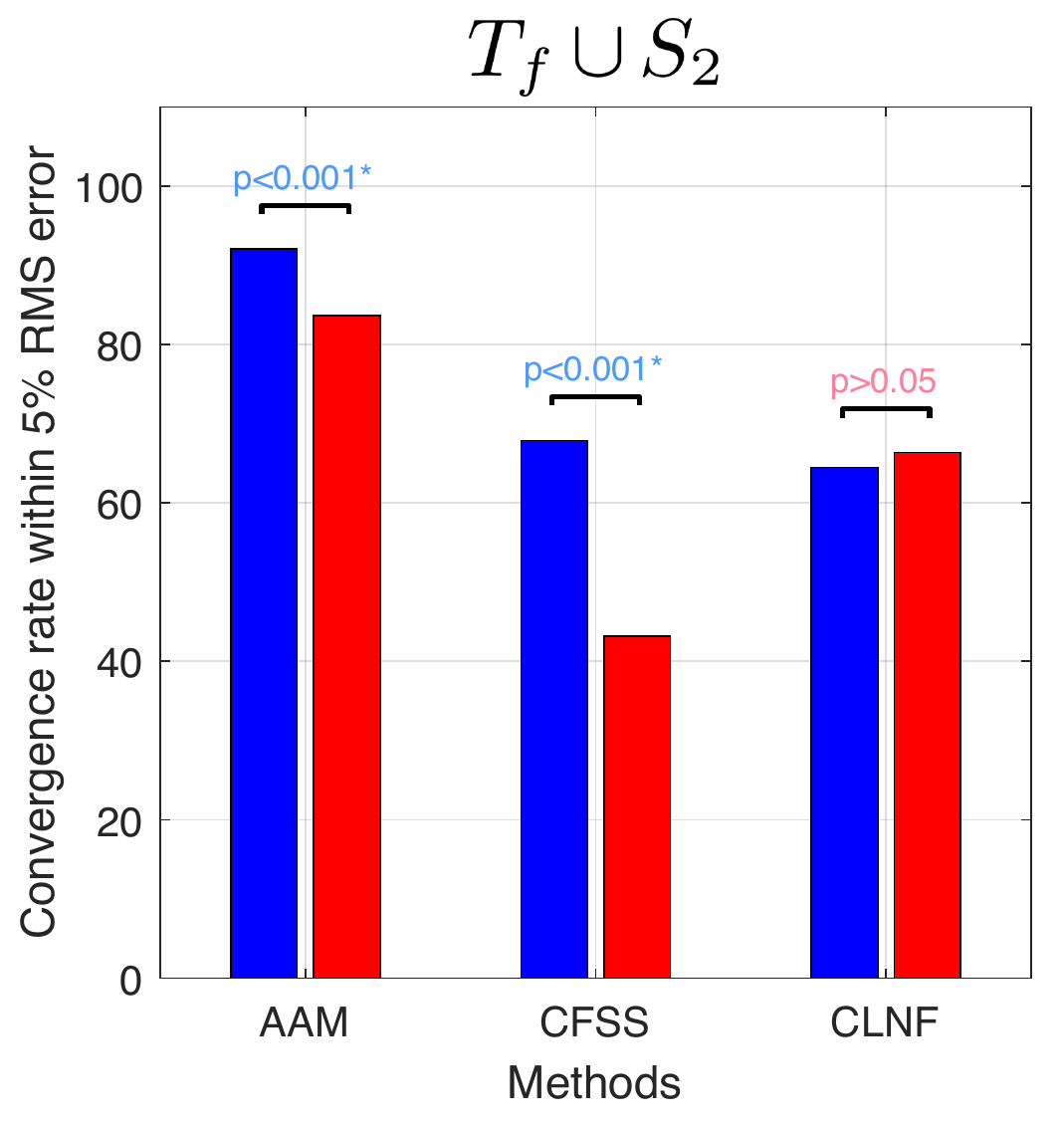}}\hspace{1mm} 
\subfigure{\includegraphics[width=.32\textwidth]{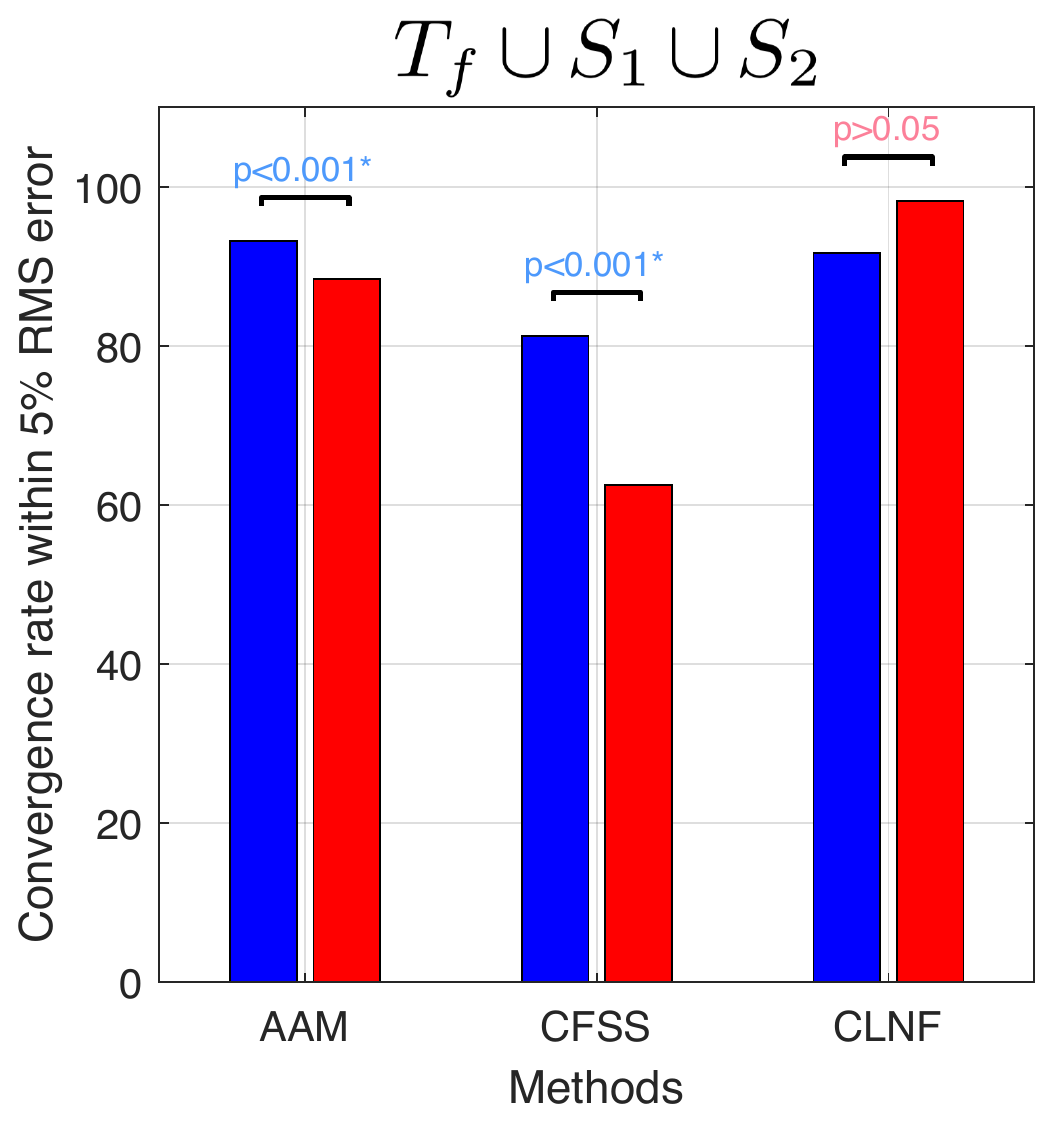}}\hspace{1mm} 
\caption{Experiment 3: Comparison of convergence percentage within 5\% tolerance of RMS fitting error obtained on healthy subset (H) and dementia subset (D) of \Tf~ using various versions of three methods AAM, CFSS, and CLNF trained on configurations $S_1, S_2, S_1 \cup S_2, T_f \cup S_1, \; T_f \cup S_2, \; T_f \cup S_1 \cup S_2$. RMS fitting errors are computed over the standard 68 landmark points (whole face). }
\label{Fig:Expt3}
\vspace{-.2cm}
\end{figure*}

\begin{figure*}[t]
\centering
\subfigure{\includegraphics[width=.32\textwidth]{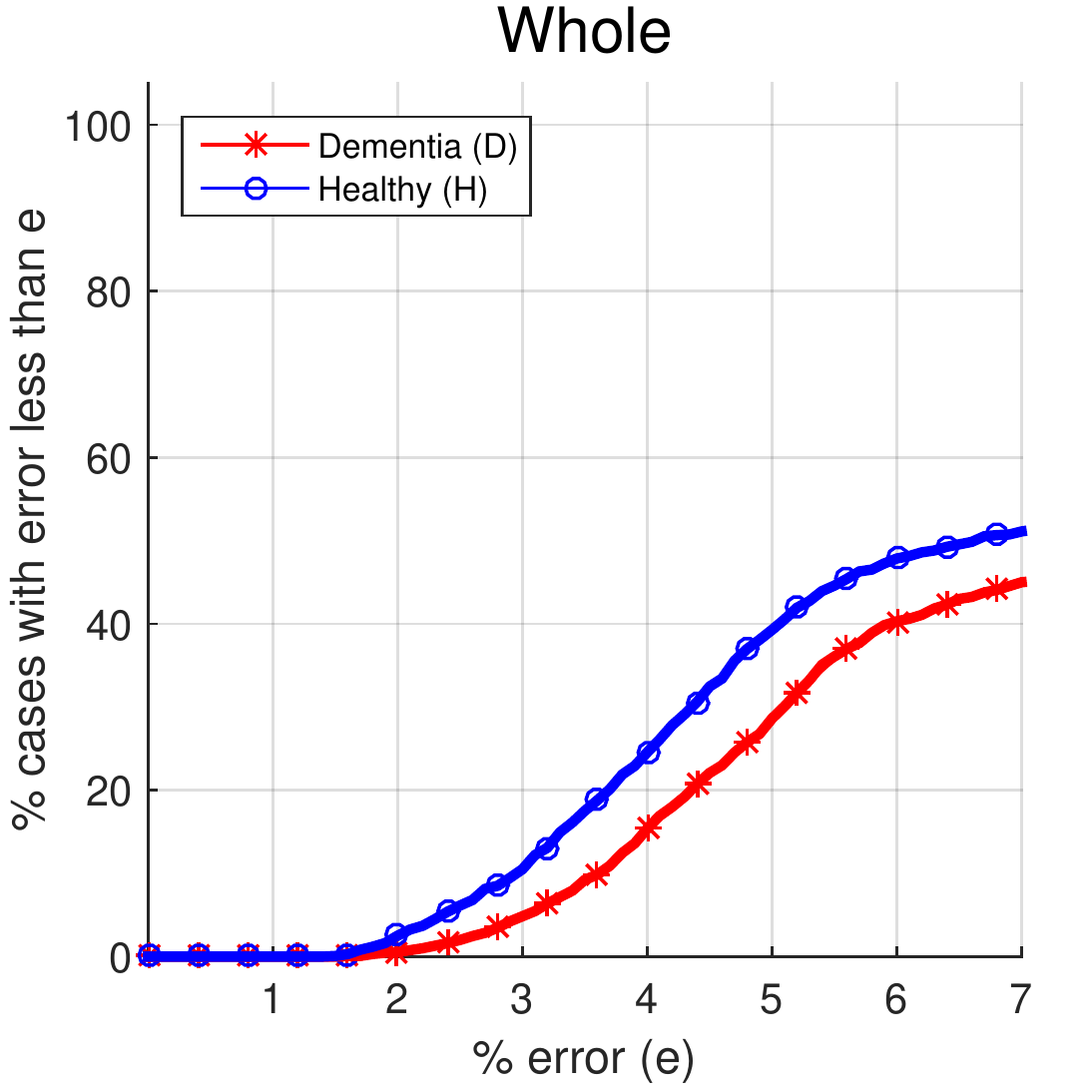}}\hspace{1mm}
\subfigure{\includegraphics[width=.32\textwidth]{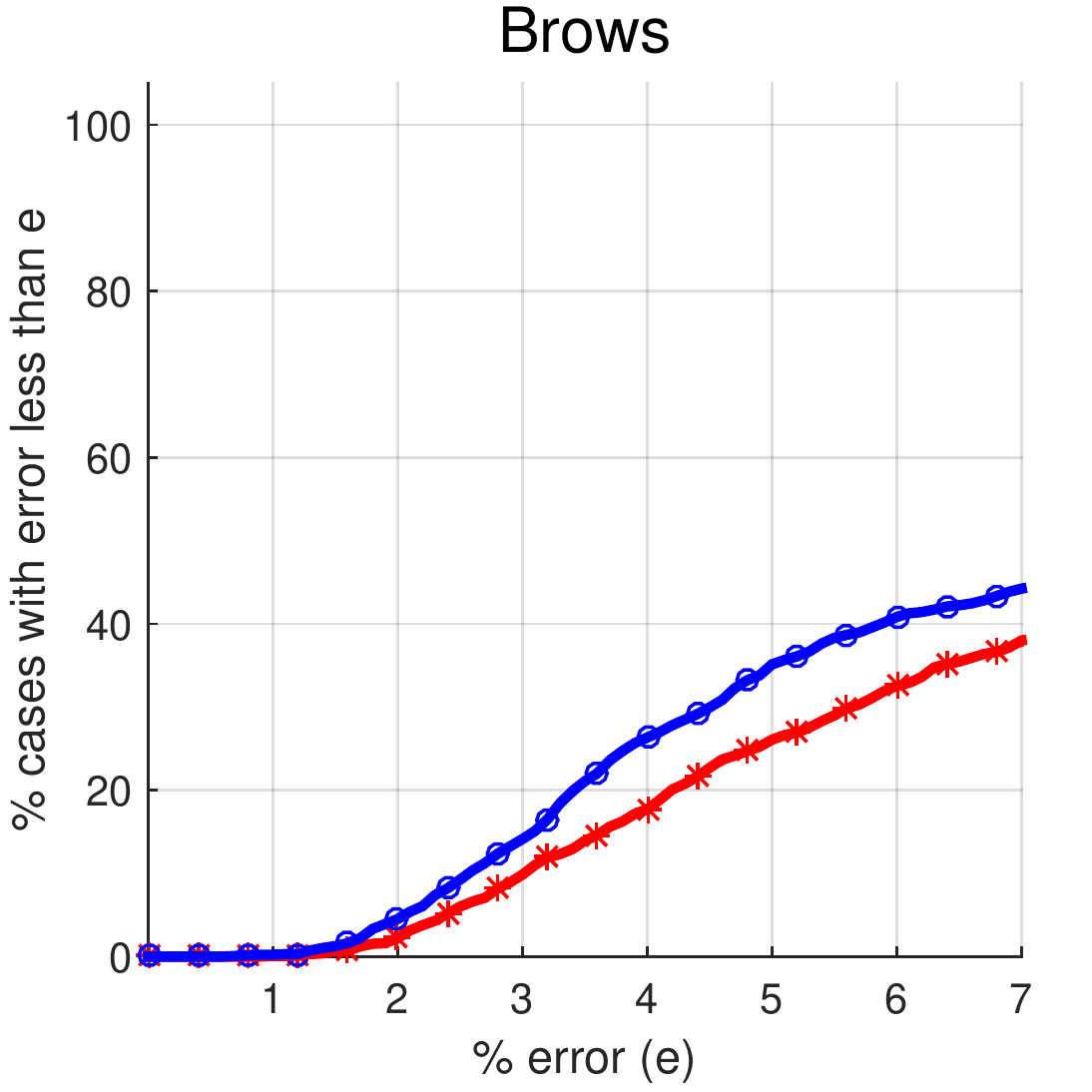}}\hspace{1mm}
\subfigure{\includegraphics[width=.32\textwidth]{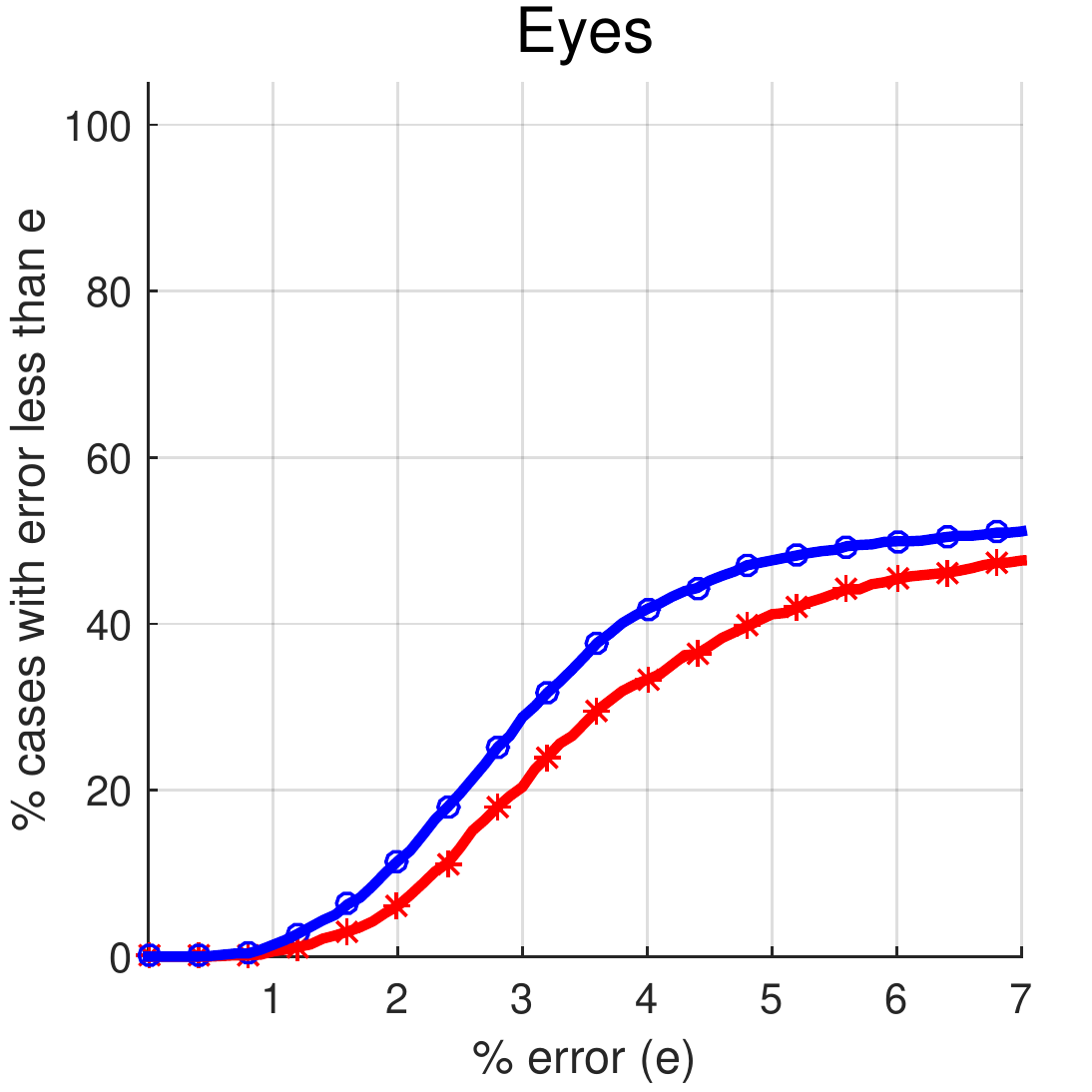}}\hspace{1mm}
\subfigure{\includegraphics[width=.32\textwidth]{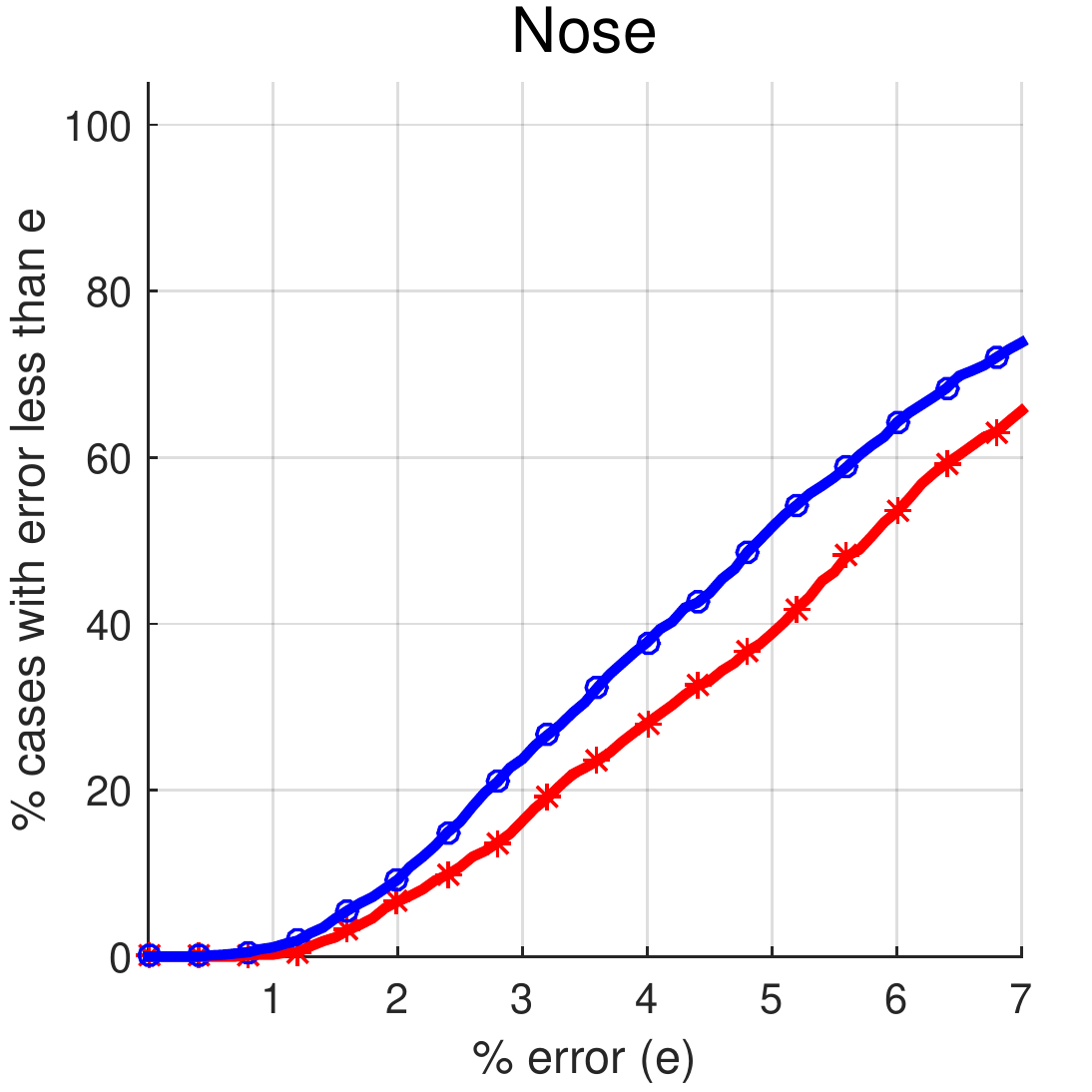}}\hspace{1mm} 
\subfigure{\includegraphics[width=.32\textwidth]{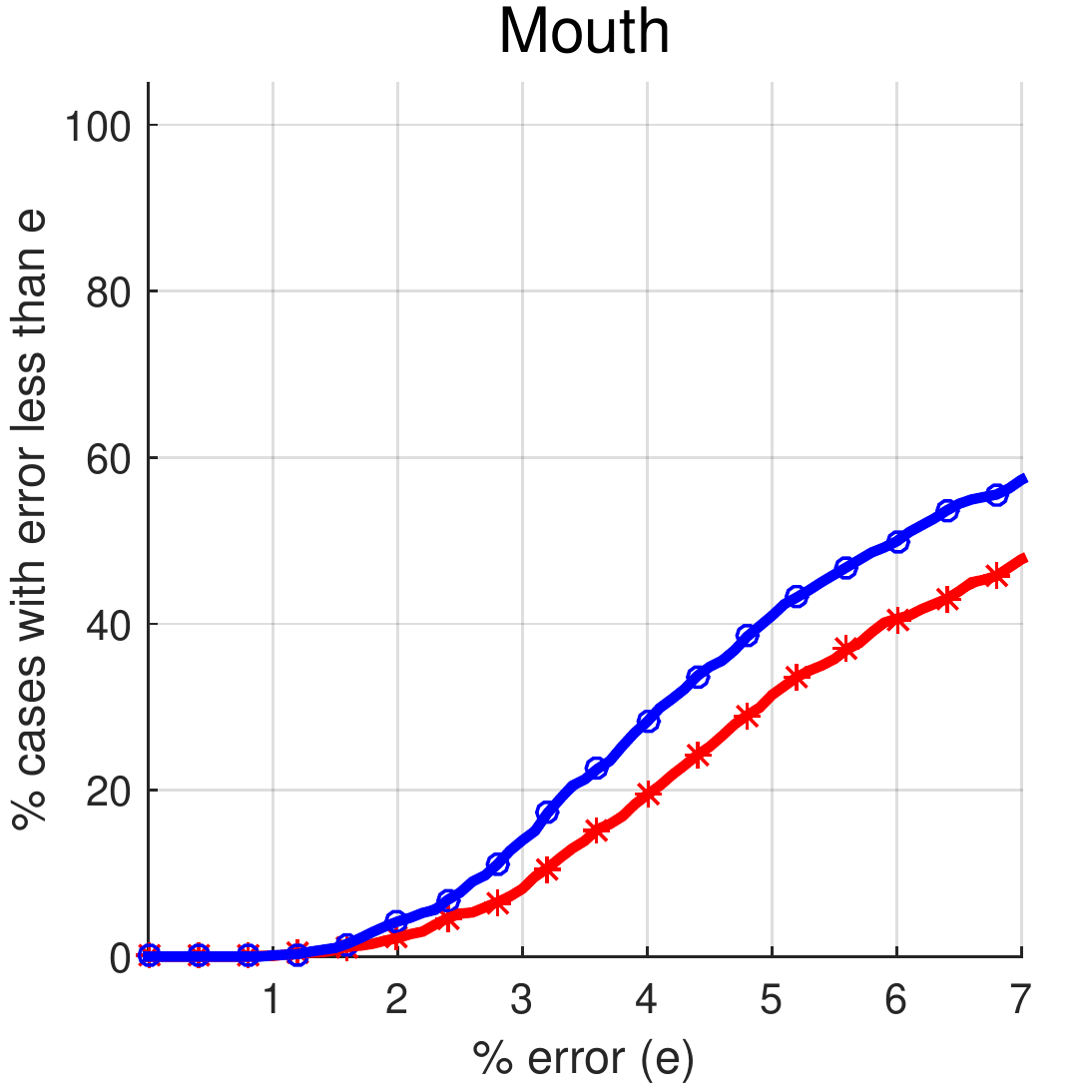}}\hspace{1mm} 
\caption{Experiment 4: Comparison of the average convergence curves obtained on healthy subset (H) and dementia subset (D) of \Tp. The values on y-axis are averaged over four methods: AAM, FAN-2D, FAN-3D, and PRNet.}
\label{Fig:Expt4}
\end{figure*}

Figures \ref{Fig:Expt1}, \ref{Fig:Expt2}, and \ref{Fig:Expt4} show the average convergence curves obtained on healthy (H) and dementia (D) subsets of \Tf\thinspace and \Tp\thinspace using different methods from experiments 1, 2 and 4 respectively. In these figures, the x-axis shows the RMS fitting error normalized by the face size (diagonal), while the y-axis shows the percentage of cases with a fitting error less than the corresponding x-axis value averaged over all methods included in the evaluation. Figure \ref{Fig:Expt3} shows the convergence rates obtained on the 68 landmark points (whole face) for healthy (H) and dementia (D) subsets of \Tf\thinspace using the re-trained versions of methods AAM, CFSS, and CLNF on the training configurations of experiment 3. 

The general trend in Experiments 1-3 show that the relationship between convergence rates of all evaluated methods and dementia is significant on frontal faces (\Tf). Although increasing the variation in the training data by including images from various datasets (\Tf, \Sf, and \Ss) improves the performance on both healthy and dementia subsets of \Tf, the difference between the convergence rates for these two subsets remains large and statistically significant (Experiments 2-3). Results of Experiment 4 show a similar trend on profile faces (\Tp) with less difference between the convergence rates obtained for healthy and dementia subsets when compared to frontal faces (\Tf). 

\begin{table*}[t]
\begin{small}
\caption{Experiment 1: Comparison of convergence percentage within 5\% tolerance of RMS fitting error obtained on healthy subset (H) and dementia subset (D) of \Tf. p-values are color coded with respect to three standard significant levels 0.05, 0.01 and 0.001.}
\begin{center}
\begin{tabu}{|cl|cc|cc|cc|cc|cc|cc|}
 \hline
 & \multirow{3}{*}{Methods} & & & & & & & & & & & & \\
 & & \multicolumn{2}{c|}{Whole} & \multicolumn{2}{c|}{Jaw} & \multicolumn{2}{c|}{Brows} & \multicolumn{2}{c|}{Nose} & \multicolumn{2}{c|}{Eyes} & \multicolumn{2}{c|}{Mouth} \\
 & & H & D & H & D & H & D & H & D & H & D & H & D\\
 \hline
& \multirow{2}{*}{CLNF}  & 71.88 & 63.39 & 24.15 & 26.49 & 50.00 & 42.26 & 84.94 & 77.38 & 85.51 & 72.32 & 84.66 & 75.60\\
\rowfont{\scriptsize}
& & \multicolumn{2}{c|}{\pval{-3}} &  \multicolumn{2}{c|}{\fail{0.367}} & \multicolumn{2}{c|}{\fail{0.078}} & \multicolumn{2}{c|}{\pval{-13}} & \multicolumn{2}{c|}{\pval{-4}} & \multicolumn{2}{c|}{\pval{-11}}\\
\hline
& \multirow{2}{*}{CFSS}   & 80.11 & 65.77 & 27.27 & 27.98 & 61.08 & 50.00 & 90.63 & 88.69 & 85.51 & 77.38 & 90.34 & 79.17\\
\rowfont{\scriptsize}
&  & \multicolumn{2}{c|}{\pval{-6}} & \multicolumn{2}{c|}{\fail{0.848}} & \multicolumn{2}{c|}{\pval{-3}} & \multicolumn{2}{c|}{\pval{-4}} & \multicolumn{2}{c|}{\pval{-8}} & \multicolumn{2}{c|}{\pval{-15}}\\
\hline
& \multirow{2}{*}{AAM}    & 87.78 & 71.73 & 44.89 & 29.76 & 62.50 & 44.94 & 95.45 & 95.24 & 94.60 & 87.80 & 93.75 & 86.31\\
\rowfont{\scriptsize}
& & \multicolumn{2}{c|}{\pval{-19}} & \multicolumn{2}{c|}{\pval{-7}} & \multicolumn{2}{c|}{\pval{-7}} & \multicolumn{2}{c|}{\pval{-6}} & \multicolumn{2}{c|}{\pval{-6}} & \multicolumn{2}{c|}{\pval{-16}}\\
\hline
& \multirow{2}{*}{FAN-2D} & 81.53 & 61.90 & 19.89 & 19.94 & 68.18 & 56.25 & 83.52 & 63.39 & 99.43 & 96.43 & 97.44 & 85.12\\
\rowfont{\scriptsize}
& & \multicolumn{2}{c|}{\pval{-9}} & \multicolumn{2}{c|}{\fail{0.095}} & \multicolumn{2}{c|}{\fail{0.065}} & \multicolumn{2}{c|}{\pval{-17}} & \multicolumn{2}{c|}{\pval{-5}} & \multicolumn{2}{c|}{\pval{-16}}\\
\hline
& \multirow{2}{*}{FAN-3D} & 71.88 & 50.60 & 18.18 & 12.20 & 61.93 & 52.98 & 85.80 & 65.48 & 98.86 & 94.64 & 98.58 & 89.88\\
\rowfont{\scriptsize}
& & \multicolumn{2}{c|}{\pval{-13}} & \multicolumn{2}{c|}{\pval{-5}} & \multicolumn{2}{c|}{\pass{0.013}} & \multicolumn{2}{c|}{\pval{-16}} & \multicolumn{2}{c|}{\pval{-5}} & \multicolumn{2}{c|}{\pval{-11}}\\
\hline
& \multirow{2}{*}{MDM}    & 91.76 & 70.24 & 36.36 & 25.30 & 65.06 & 47.02 & 97.16 & 90.18 & 97.44 & 91.96 & 95.74 & 85.42\\
\rowfont{\scriptsize}
& & \multicolumn{2}{c|}{\pval{-17}} & \multicolumn{2}{c|}{\pval{-6}} & \multicolumn{2}{c|}{\pval{-12}} & \multicolumn{2}{c|}{\pval{-4}} & \multicolumn{2}{c|}{\pval{-7}} & \multicolumn{2}{c|}{\pval{-10}}\\
\hline
& \multirow{2}{*}{PRNet}  & 76.14 & 64.29 & 20.74 & 18.75 & 72.73 & 65.18 & 93.75 & 88.10 & 95.45 & 89.88 & 86.08 & 76.49\\
\rowfont{\scriptsize}
& & \multicolumn{2}{c|}{\pval{-7}} & \multicolumn{2}{c|}{\pas{0.008}} & \multicolumn{2}{c|}{\pas{0.004}} & \multicolumn{2}{c|}{\pval{-12}} & \multicolumn{2}{c|}{\pval{-4}} & \multicolumn{2}{c|}{\pval{-4}}\\
\hline
\end{tabu}
\label{Tab:Expt1}
\end{center}
\end{small}
\end{table*}

\begin{table*}[t]
\begin{small}
\caption{Experiment 2: Comparison of convergence percentage within 5\% tolerance of RMS fitting error obtained on healthy subset (H) and dementia subset (D) of \Tf. p-values are color coded with respect to three standard significant levels 0.05, 0.01 and 0.001.}
\begin{center}
\begin{tabu}{|cl|cc|cc|cc|cc|cc|cc|}
 \hline
 & \multirow{3}{*}{Methods} & & & & & & & & & & & & \\
 & & \multicolumn{2}{c|}{Whole} & \multicolumn{2}{c|}{Jaw} & \multicolumn{2}{c|}{Brows} & \multicolumn{2}{c|}{Nose} & \multicolumn{2}{c|}{Eyes} & \multicolumn{2}{c|}{Mouth} \\
 & & H & D & H & D & H & D & H & D & H & D & H & D\\
 \hline
 & \multirow{2}{*}{CLNF} & 87.78 & 75.89 & 65.91 & 48.51 & 75.57 & 63.39 & 90.63 & 77.68 & 91.19 & 81.55 & 91.19 & 80.65 \\
 \rowfont{\scriptsize}
& & \multicolumn{2}{c|}{\pval{-18}} &  \multicolumn{2}{c|}{\pval{-8}} & \multicolumn{2}{c|}{\pval{-5}} & \multicolumn{2}{c|}{\pval{-19}} & \multicolumn{2}{c|}{\pval{-9}} & \multicolumn{2}{c|}{\pval{-18}}\\
\hline
 & \multirow{2}{*}{CFSS} & 56.53 & 36.9 & 24.43 & 13.1 & 34.09 & 28.87 & 82.67 & 70.83 & 65.91 & 57.14 & 87.78 & 65.48 \\
 \rowfont{\scriptsize}
& & \multicolumn{2}{c|}{\pval{-9}} &  \multicolumn{2}{c|}{\pval{-10}} & \multicolumn{2}{c|}{\pass{0.018}} & \multicolumn{2}{c|}{\pval{-10}} & \multicolumn{2}{c|}{\pval{-4}} & \multicolumn{2}{c|}{\pval{-22}}\\
\hline
 & \multirow{2}{*}{AAM} & 94.32 & 83.93 & 62.22 & 48.51 & 69.89 & 59.82 & 98.86 & 92.86 & 98.01 & 91.37 & 97.16 & 84.52 \\
 \rowfont{\scriptsize}
& & \multicolumn{2}{c|}{\pval{-17}} &  \multicolumn{2}{c|}{\pval{-6}} & \multicolumn{2}{c|}{\pas{0.004}} & \multicolumn{2}{c|}{\pval{-11}} & \multicolumn{2}{c|}{\pval{-7}} & \multicolumn{2}{c|}{\pval{-19}}\\
\hline
 & \multirow{2}{*}{FFAN-HG} & 96.02 & 90.18 & 70.74 & 63.99 & 67.61 & 63.1 & 98.01 & 97.02 & 94.89 & 87.5 & 98.3 & 93.45 \\
 \rowfont{\scriptsize}
& & \multicolumn{2}{c|}{\pval{-7}} &  \multicolumn{2}{c|}{\pas{0.003}} & \multicolumn{2}{c|}{\pas{0.003}} & \multicolumn{2}{c|}{\pval{-3}} & \multicolumn{2}{c|}{\pval{-3}} & \multicolumn{2}{c|}{\pval{-7}}\\
\hline
\end{tabu}
\label{Tab:Expt2}
\end{center}
\end{small}
\vspace{-.4cm}
\end{table*}

From Experiment 1 (Table~\ref{Tab:Expt1} and Figure~\ref{Fig:Expt1}), we can see that the difference in convergence rates obtained on the whole face between healthy and dementia subsets of \Tf\thinspace is large and statistically significant for every one of the seven methods evaluated. Figure~\ref{Fig:Expt1} shows that for all regions of the face the convergence curves for the healthy subset lie above the convergence curves for the dementia subset. We also notice that the difference between convergence curves is larger in the mouth, eyes, and nose regions of the faces as compared to the brows and the jaw.  This has implications in applications where the tracking of the mouth, eyes, or nose regions is important, e.g., in the detection of pain~\cite{prkachin2008structure}.

Table \ref{Tab:Expt2} and Figure \ref{Fig:Expt2} show the performance of re-trained/fine-tuned versions of methods CLNF, CFSS, AAM and FFAN-HG with images from \Tf\thinspace on various regions of the face. Comparing the results reported in Tables \ref{Tab:Expt2} and \ref{Tab:Expt1}, we see that the performance for all methods except CFSS has largely increased on both healthy and dementia subsets after including images from \Tf\thinspace in the training data. This is possibly because of the searching mechanism used in the CFSS model and the significant difference between the size of \Tf\thinspace and the data used originally to train it.

Although we see a boost in the convergence curves for most regions when comparing Figure \ref{Fig:Expt2} to Figure \ref{Fig:Expt1}, the convergence rates for the dementia subset are still lower compared to those for the healthy subset of \Tf and the difference is significant for all regions of the face (Table \ref{Tab:Expt2}). This trend is particularly noticeable in the jaw and in the eyes.

Figure \ref{Fig:Expt3} shows the convergence rates obtained on healthy (H) and dementia (D) subsets of \Tf\thinspace using the re-trained versions of methods AAM, CFSS and CLNF on the following training configurations: $S_1, S_2, S_1 \cup S_2, T_f \cup S_1, \; T_f \cup S_2, \; T_f \cup S_1 \cup S_2$. We see that the convergence rates for healthy and dementia subsets vary largely by configuration; however, the difference between them remains significant for all configurations and methods (except for method CLNF when trained on $T_f\cup S_2$ and $T_f \cup S_1 \cup S_2$). A similar trend was also observed in the convergence rates for all regions of the face (included in the supplementary materials). Comparing the results of Experiments 1, 2 and 3, we notice that the inclusion of additional variation in the training data can help to improve the performance in general, but it does not help with mitigating the gap between the performance on healthy and dementia subsets.

Table \ref{Tab:Expt4} and Figure \ref{Fig:Expt4} show the performance of AAM, FAN-2D, FAN-3D, and PRNet when evaluated on the profile face of \Tp. Performance is poor as compared to performance on \Tf; but, similar to the previous experiments, we see that the average convergence curves for the healthy subset lie above the curves for the dementia subset in all regions of the face. The difference between the performance on healthy and dementia subsets of \Tp\thinspace is smaller compared to the ones for frontal faces in \Tf, yet it is significant on some regions of the face such as the nose and the mouth. 

\begin{table*}[t]
\begin{small}
\caption{Experiment 4: Comparison of convergence percentage within 5\% tolerance of RMS fitting error obtained on healthy subset (H) and dementia subset (D) of \Tp. p-values are color coded with respect to three standard significant levels 0.05, 0.01 and 0.001.}
\begin{center}
\begin{tabu}{|cl|cc|cc|cc|cc|cc|}
 \hline
 & \multirow{3}{*}{Methods} & & & & & & & & & & \\
 & & \multicolumn{2}{c|}{Whole} & \multicolumn{2}{c|}{Brows} & \multicolumn{2}{c|}{Nose} & \multicolumn{2}{c|}{Eyes} & \multicolumn{2}{c|}{Mouth} \\
 & & H & D & H & D & H & D & H & D & H & D\\
 \hline
& \multirow{2}{*}{AAM} & 44.67 & 37.23 & 32.84 & 21.54 & 51.18 & 46.15 & 47.93 & 44 & 52.66 & 37.54 \\
\rowfont{\scriptsize} & & \multicolumn{2}{c|}{\pval{-3}} &  \multicolumn{2}{c|}{\pass{0.015}} & \multicolumn{2}{c|}{\pass{0.034}} & \multicolumn{2}{c|}{\pas{0.010}} & \multicolumn{2}{c|}{\pval{-3}}\\ \hline
 & \multirow{2}{*}{FAN-2D} & 35.21 & 26.15 & 39.64 & 32.31 & 35.21 & 21.85 & 49.41 & 44 & 35.5 & 29.23 \\
\rowfont{\scriptsize} & & \multicolumn{2}{c|}{\fail{0.193}} &  \multicolumn{2}{c|}{\fail{0.817}} & \multicolumn{2}{c|}{\pval{-3}} & \multicolumn{2}{c|}{\fail{0.268}} & \multicolumn{2}{c|}{\fail{0.173}}\\ \hline
 & \multirow{2}{*}{FAN-3D} & 36.09 & 22.15 & 36.39 & 28.31 & 43.2 & 23.08 & 49.41 & 41.23 & 38.76 & 29.23 \\
\rowfont{\scriptsize} & & \multicolumn{2}{c|}{\pass{0.043}} &  \multicolumn{2}{c|}{\fail{0.709}} & \multicolumn{2}{c|}{\pval{-6}} & \multicolumn{2}{c|}{\fail{0.143}} & \multicolumn{2}{c|}{\pass{0.030}}\\ \hline
 & \multirow{2}{*}{PRNet} & 41.12 & 28.92 & 31.66 & 22.15 & 76.92 & 64.31 & 43.79 & 35.38 & 36.98 & 29.85 \\
\rowfont{\scriptsize} & & \multicolumn{2}{c|}{\fail{0.192}} &  \multicolumn{2}{c|}{\fail{0.951}} & \multicolumn{2}{c|}{\pval{-4}} & \multicolumn{2}{c|}{\fail{0.553}} & \multicolumn{2}{c|}{\fail{0.082}}\\ \hline
\end{tabu}
\label{Tab:Expt4}
\end{center}
\end{small}
\vspace{-.2cm}
\end{table*}

\section{Conclusions}
\label{sec:conclusions}
Accurate detection of facial landmark points is an important requirement for a wide range of clinical applications involving older adults and/or individuals with a cognitive or a physical disability. In this paper, we provide a comprehensive evaluation of state-of-the-art facial landmark detection on faces of older adults with and without dementia. Our evaluation demonstrates an algorithmic bias in state-of-the-art facial landmark detection methods, which affects the performance of these methods for older adults with dementia. Furthermore, our empirical analysis shows that techniques such as fine-tuning and re-training can improve the performance for both groups; however, these methods cannot reduce the gap between the performance for adults with and without dementia. As interest in employing facial analysis methods in clinical applications grows~\cite{asgarian2018hybrid, ashraf2016automated}, our study sheds light on the limitations of existing facial landmark detection methods and the challenges of applying these methods to clinical populations. In future work, we plan to investigate potential solutions to overcome these biases in facial landmark detection methods.

{\small
\bibliographystyle{ieee}
\bibliography{egbib}
}

% Adding the supplementary materials
\includepdf[pages=1]{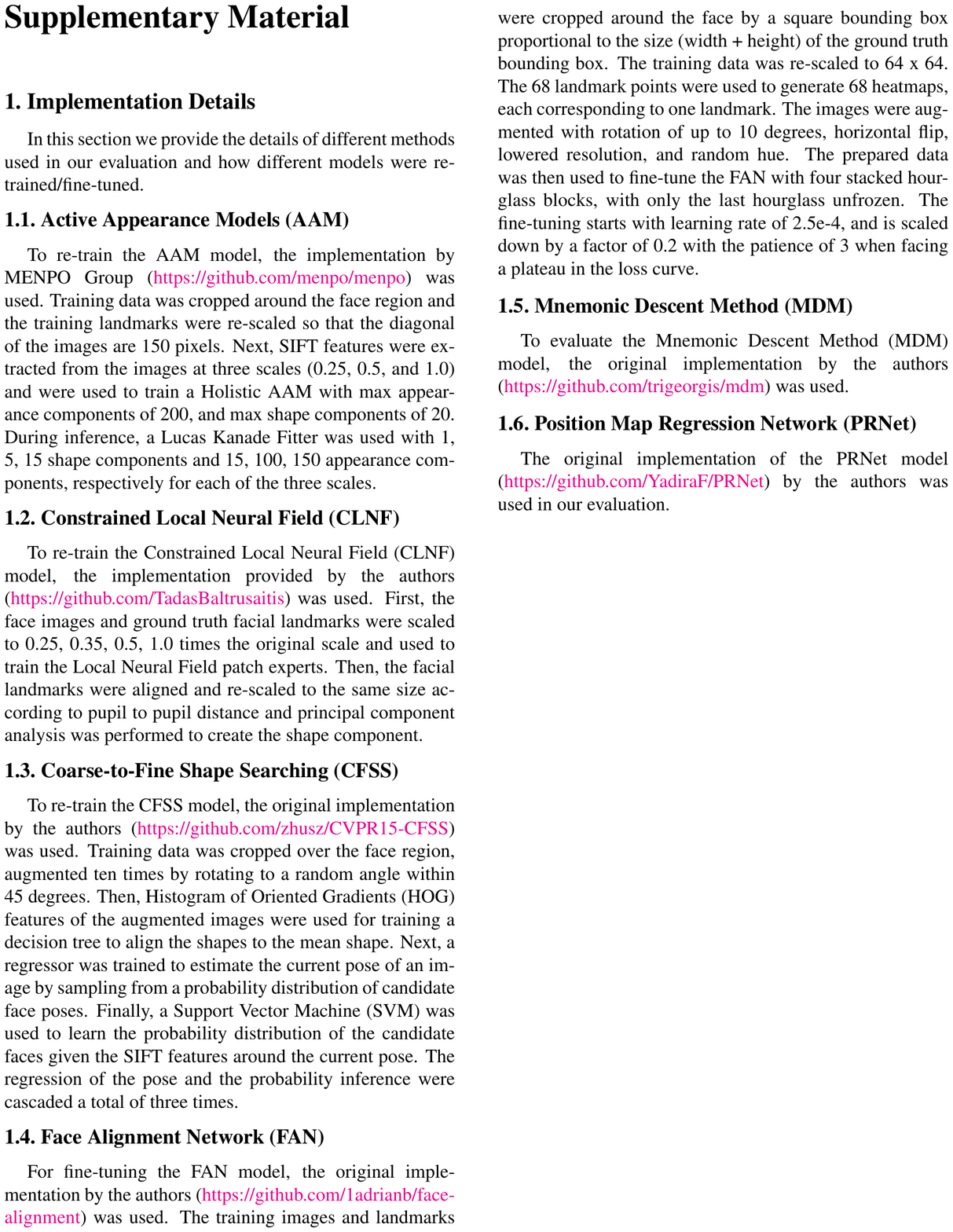}
\includepdf[pages=2]{Supplementary_Materials.pdf}
\includepdf[pages=3]{Supplementary_Materials.pdf}
\includepdf[pages=4]{Supplementary_Materials.pdf}
\includepdf[pages=5]{Supplementary_Materials.pdf}

\end{document}